\documentclass[pdflatex,sn-nature]{sn-jnl}
\catcode`\|=12\relax
\usepackage[T1]{fontenc}
\let\orcidlogo\relax
\usepackage{orcidlink}
\usepackage{float}
\usepackage{placeins}
\usepackage{anyfontsize}
\usepackage{siunitx}
\usepackage{graphicx}
\usepackage{multirow}
\usepackage{amsmath,amssymb,amsfonts}
\usepackage{amsthm}
\usepackage{mathrsfs}
\usepackage[title]{appendix}
\usepackage{xcolor}
\usepackage{textcomp}
\usepackage{manyfoot}
\usepackage{booktabs}
\usepackage{url}
\usepackage{longtable}
\usepackage{tabularx}
\usepackage{array}
\usepackage{calc}
\providecommand{\real}[1]{#1}
\usepackage{bibunits}
\makeatletter
\newcounter{savebibitem}
\let\oldthebibliography\thebibliography
\renewcommand{\thebibliography}[1]{%
  \oldthebibliography{#1}%
  \ifnum\value{savebibitem}>0
    \setcounter{NAT@ctr}{\value{savebibitem}}%
  \fi
}
\let\oldendthebibliography\endthebibliography
\renewcommand{\endthebibliography}{%
  \setcounter{savebibitem}{\value{NAT@ctr}}%
  \oldendthebibliography%
}
\makeatother
\usepackage{geometry}
\begin{document}

\title[Individual-level interventions against sycophantic AI reduce its appeal but not its persuasiveness]{\textsc{Individual-level interventions against sycophantic AI reduce its appeal but not its persuasiveness}}

\author*[1]{\fnm{Meryl} \sur{Ye}}\email{merylye@cmu.edu}
\author[1]{\fnm{Robert} \sur{Kraut}}
\author[1,2]{\fnm{Steve} \sur{Rathje}}

\affil[1]{\orgname{Carnegie Mellon University}}
\affil[2]{\orgname{New York University}}

\begin{bibunit}[naturemag]

\abstract{AI chatbots can be ``sycophantic,'' or overly agreeable and flattering toward users. Sycophantic AI has been shown to entrench attitudes, yet users frequently fail to recognize it (a phenomenon we call ``sycophancy blindness''). We tested whether increasing users' awareness of sycophancy protects them from its harmful effects in two preregistered experiments (\textit{n} = 1,590). In the first, participants received a brief written warning about sycophancy before conversing with a sycophantic chatbot. In the second, participants watched a video of a sycophantic AI validating several other users, including users on opposite sides of the same conflict, before interacting with it themselves. Both interventions changed how participants evaluated the AI. The warning reduced the AI's perceived objectivity, and the video reduced enjoyment of the AI --- an effect mediated by the reduced belief that its validation was uniquely earned. We then pooled our experiments with two prior studies of sycophancy awareness interventions (six interventions total, \textit{n} = 3,982). The pattern across experiments was consistent: while the interventions made the sycophantic AI appear less objective and trustworthy,  none reduced its persuasiveness. These results suggest that individual-level interventions, such as warning labels or AI literacy, may not be enough to protect users from AI harms.}

\maketitle
\newpage
\section*{Introduction}\label{sec:intro}
There has been rising concern among academics, policymakers, and technology companies that AI models exhibit ``sycophancy,'' a family of behaviors in which AI systems excessively agree with, flatter, or validate users \cite{perez2023discovering,sharma2024towards,ye2026counts}. For instance, a recent study found that AI chatbots validate users approximately 50\% more often than humans do \cite{cheng2025elephant}, and even brief interactions with sycophantic AI entrench users' pre-existing attitudes and reduce their willingness to repair interpersonal conflicts \cite{rathje2025sycophantic,cheng2026prosocial}. These findings are concerning because sycophantic AI leaves users more extreme and more confident in contested positions and less inclined to take steps toward resolving conflicts \cite{rathje2025sycophantic,cheng2026prosocial}. Research on biased assimilation suggests that these changes in beliefs and behavior contribute to attitude polarization and prolonged disagreement \cite{lord1979biased}. Sycophancy has also entered public discourse, most notably spiking in April 2025 when OpenAI rolled back an update to GPT-4o after widespread public complaints that the model had become sycophantic \cite{openai2025sycophancy}.

A relevant question is whether individual-level interventions can protect users from the harmful effects of certain AI behaviors, including (but not limited to) sycophancy. By individual-level interventions, we refer to measures that directly face or target the user, such as warnings, disclosures, and educational materials that teach people to recognize a behavior, as opposed to model-level changes that alter how the system behaves in the first place. The question has practical importance: AI companies are already implementing such interventions around their products, and educators are adopting AI literacy curricula on the assumption that a user who recognizes a problematic behavior will be less affected by it.

Yet, even as awareness of sycophancy has grown, users often fail to notice sycophancy in their own conversations with AI. One study found that participants who interacted with sycophantic chatbots rated these chatbots as unbiased even though third-party annotators judged them as biased \cite{rathje2025sycophantic}. In another study, novices debugged machine learning models with either a highly sycophantic or a less sycophantic chatbot \cite{10.1145/3772318.3791365}. The sycophantic chatbot validated participants' misconceptions rather than correcting them and harmed their debugging performance (participants improved their models by 5\% on average, versus 49\% with the less sycophantic chatbot), yet 71\% of participants did not notice the sycophancy and rated the two chatbots as similarly helpful and reliable. This ``sycophancy blindness'' mirrors findings from past work showing that in social interactions people uncritically accept excessive flattery \cite{gordon1996impact,Usman2024} and ideas that align with their prior beliefs \cite{lord1979biased,ross2013naive}.

We propose that sycophancy is appealing in part because people attribute an AI chatbot's validation to the quality of their own ideas or personal characteristics as opposed to the chatbot's tendency to validate anyone. A validator is \emph{selective} when its response depends on the quality of what it evaluates, and \emph{unselective} when it affirms independent of quality. Only a selective validator's agreement is meaningful. A similar dynamic appears in romantic attraction. People who expressed interest in many of their speed-dating partners were desired less in return, because they came across as interested in everyone rather than any one person \cite{eastwick2007selective,kenny1988interpersonal}. In both cases, the value of positive feedback depends on whether it reflects genuine selectivity. If sycophancy blindness enables its influence, then confronting it, whether by warning users about sycophancy or making the AI's unselective validation salient, should reduce its appeal. 

We report two original preregistered experiments testing this thesis. In Study 1, participants received a brief written warning about sycophancy before talking with a sycophantic AI chatbot. In Study 2, participants instead watched the sycophantic AI validate other users before interacting with it. To situate our results within the growing body of evidence, we then pooled our experiments with four interventions from two other recent studies \cite{ibrahim2026warninglabelsshiftperceptions,marvel2026inoculating}, six individual-level interventions in total (\textit{n} = 3,982). Across both experiments and the pooled analysis, the interventions changed how participants perceived the sycophantic AI without reducing its persuasive influence.

\section*{Results}\label{sec:results}

Throughout the paper, we distinguish two types of outcomes. \textit{Perceptions} of the AI are participants' judgments of the chatbot itself, including its objectivity, trustworthiness, and how enjoyable it was. \textit{Topic attitudes} are participants' positions on the issue they discussed with it, measured as attitude certainty and extremity. We use \textit{persuasive influence} to refer to the reinforcement of topic attitudes, operationalized as changes in attitude certainty and extremity or post-conversation judgments of rightness. We focus on whether interventions that change perceptions of the AI also reduce its influence on topic attitudes.

\subsection*{Study 1: The Effects of a Brief Written Warning About Sycophancy}

\begin{figure}[t]
\centering
\includegraphics[width=0.9\linewidth]{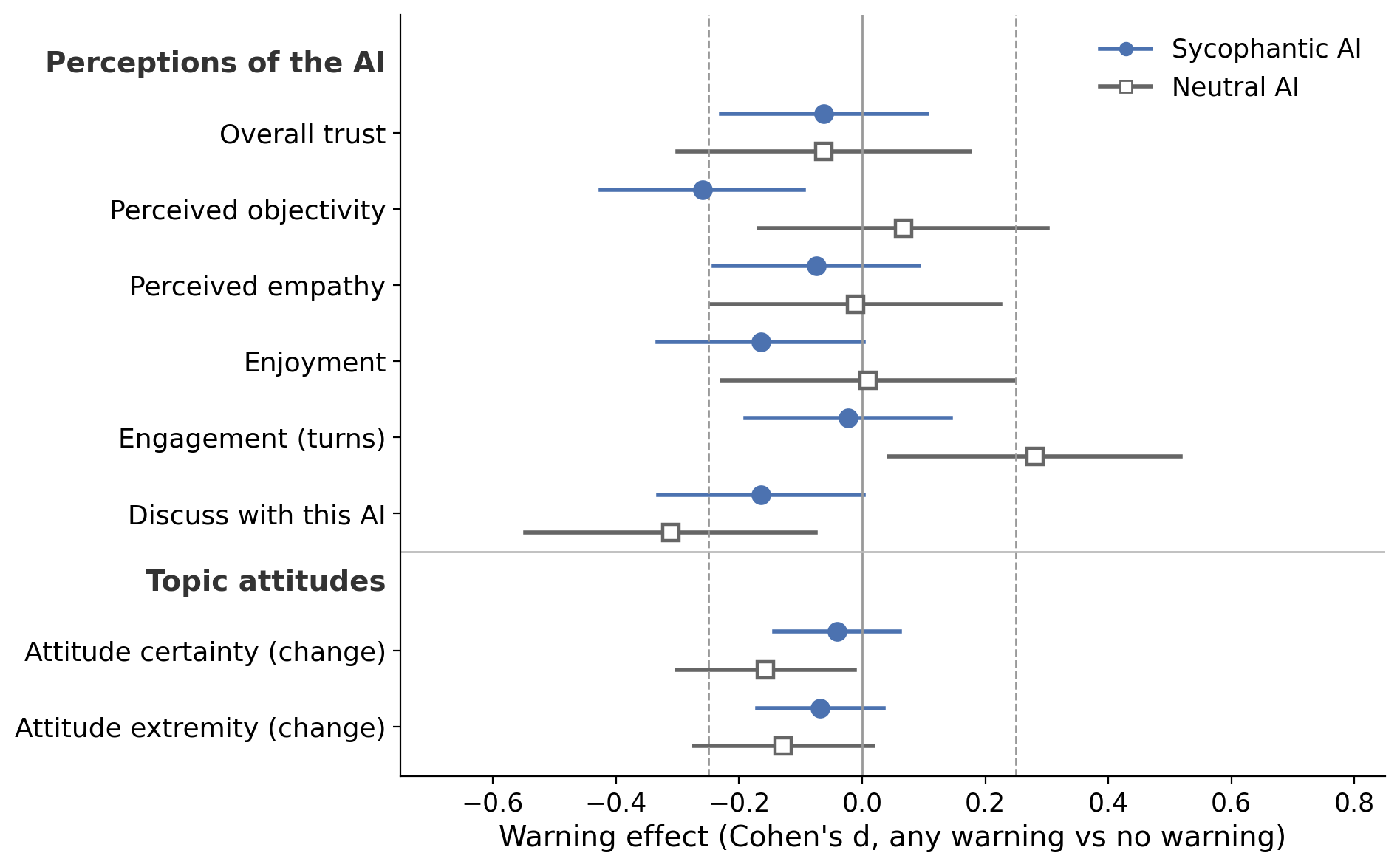}
\caption{Effects of receiving a warning, within the sycophantic AI conditions (blue circles) and within the neutral AI condition (open grey squares). Study 1 crossed three chatbot types (flattery-oriented, agreement-oriented, neutral) with three warning conditions (a warning about flattery, a warning about agreement, or no warning). Because the two wordings of each warning were preregistered as a single collapsed factor, and the two warning types did not reliably differ in their effects (all matched-versus-mismatched \textbar d\textbar\ $< 0.12$; Supplementary Note~S1), each point estimates the contrast between participants who received any warning and those who received none, estimated as a simple effect in the collapsed 2 $\times$ 2 model; Supplementary Table~\ref{stab:6} reports the sycophantic-condition values. Blue circles additionally collapse across the two sycophantic chatbots, and each blue circle is plotted directly above its paired neutral-condition square. Effects are Cohen's \emph{d} with 95\% CIs; dashed lines mark equivalence bounds at $\pm 0.25$. Contrasts within the sycophantic conditions are Holm-corrected; contrasts within the neutral condition are exploratory and uncorrected.}\label{fig:1}
\end{figure}

In Study 1, 940 participants were randomly assigned to discuss either a political topic or a recent personal conflict with one of three chatbots that differed in their sycophantic behavior: a flattery-oriented chatbot that praised the user's personal qualities, an agreement-oriented chatbot that treated the user's position as correct and supported it with agreeing arguments, or a neutral chatbot that presented multiple perspectives. Before the conversation, participants received either a brief warning that chatbots can be flattering, a warning that chatbots can be agreeable, or no warning at all.

Flattery and agreement are conceptually distinct forms of sycophancy, one directed at the person and one at their position \cite{ye2026counts}, so a warning describing the specific behavior a user is about to encounter could plausibly be more protective than a warning describing a different behavior.
We preregistered two hypotheses: (H1) warnings would reduce trust, enjoyment, and engagement, and (H2) warnings matched to the AI’s behavior would produce larger reductions in trust, enjoyment, and engagement than mismatched warnings. Neither was supported (H1: all \textbar d\textbar\ $\leq 0.17$, all $p \geq .37$; H2: all \textbar d\textbar\ $< 0.12$, all $p = 1.00$). The H2 null is notable in that it shows even a warning tailored to the exact behavior the AI displayed conferred no additional protection. See Supplementary Note~S1 for the full results. 

We treat Study 1 as exploratory and focus on how the warning affected participants' perceptions of the AI and their topic attitudes. The agreement AI increased attitude certainty relative to the neutral AI ($d = 0.16$, $p = .004$), consistent with prior work showing that sycophantic AI shifts user topic attitudes \cite{rathje2025sycophantic}.

\paragraph{Warnings reduced perceived objectivity without reducing influence on topic attitudes.} To test whether warnings reduced the influence of sycophantic AI, we collapsed across the two sycophantic AI types and compared participants who received any warning with those who received no warning. The warning reduced perceived objectivity ($d = -0.26$, $p = .017$). However, it did not reduce the AI's influence on attitude certainty or extremity. Overall trust, perceived empathy, enjoyment, engagement, and willingness to discuss the issue with the AI were also unaffected (Fig.~\ref{fig:1}). The warnings' effect on perceived objectivity was specific to the sycophantic AI. The same contrast did not change perceived objectivity within the neutral condition ($d = 0.07$, $p = .58$), and the interaction was reliable ($d = -0.33$, $p = .026$).

\subsection*{Study 2: The Effects of Observing Others Interact with Sycophantic AI }

In Study 2, 650 participants (330 control, 320 intervention) discussed a personal conflict with a sycophantic AI that consistently affirmed the participant's perspective and expressed understanding and validation, without presenting counterarguments. Before the conversation, participants in the intervention condition watched a five-minute video showing the same sycophantic AI interacting with four other users, two pairs holding opposite positions in the same conflict. In one conflict, the two users occupied opposite roles: one was frustrated that their partner kept canceling their dates, and the other was the one doing the canceling, whose partner was upset with them. The AI validated both, siding with the canceled-on partner in the first case and with the canceling partner in the second. In the other pair, it validated both a user frustrated that a friend had not repaid a loan and a user who had borrowed from a friend and felt pressured by their repeated requests for repayment. In each case the AI affirmed whichever side the user presented, making its indiscriminate validation directly observable. After the conversation, all participants received a single response from a separate neutral AI that presented multiple perspectives on their conflict. Participants rated the sycophantic conversation AI and this neutral AI on the same perception items.
Across conditions, attitude certainty and extremity increased from before to after the conversation with the sycophantic AI (within-person $d_z$ = 0.25 and 0.24, both \emph{p} $<$ .001), consistent with prior evidence that sycophantic AI can reinforce users' topic attitudes \cite{rathje2025sycophantic}.

\begin{figure}[t]
\centering
\includegraphics[width=0.9\linewidth]{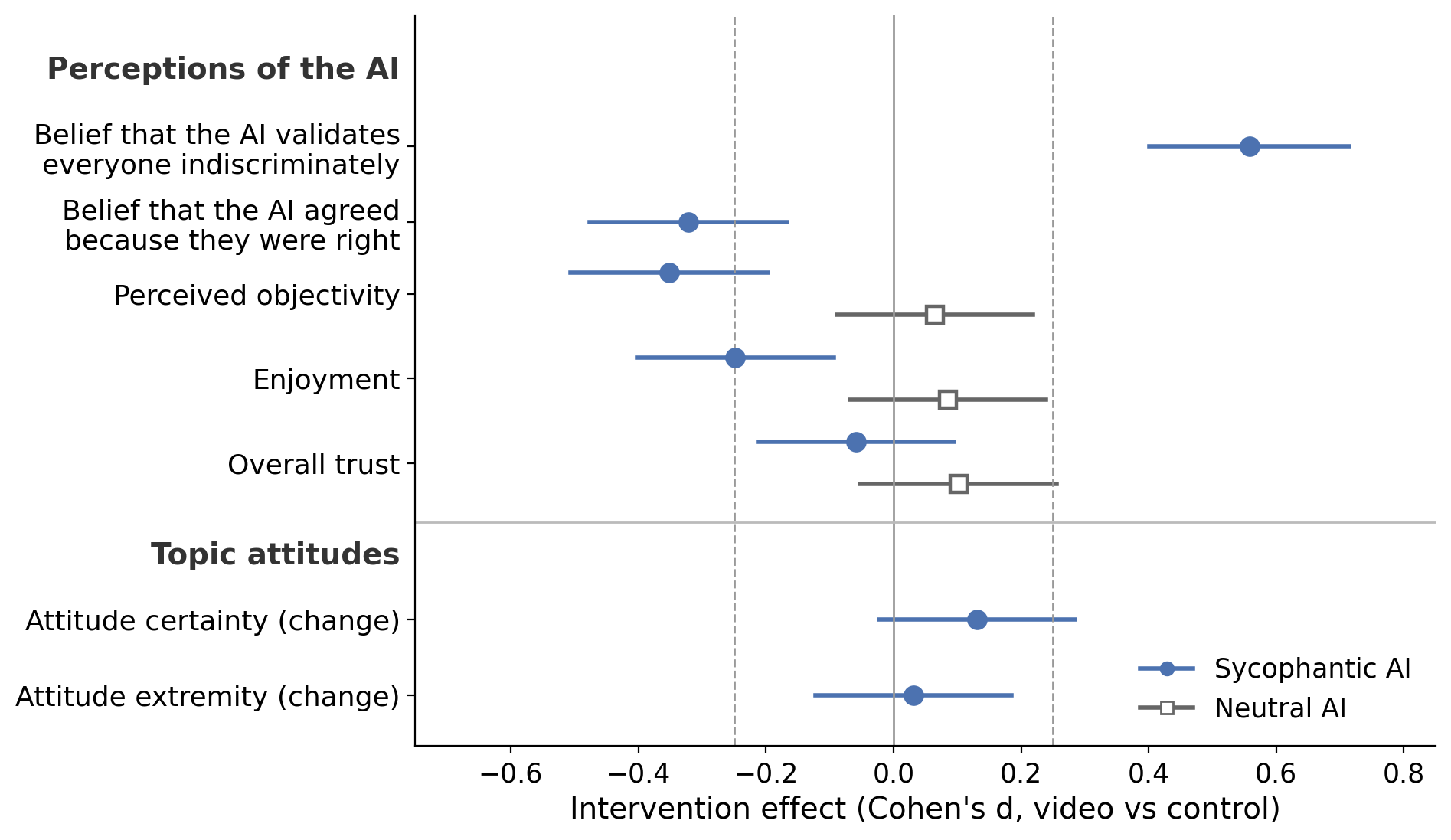}
\caption{The seven primary outcomes of Study 2 with 95\% CIs and equivalence bounds at $\pm 0.25$ (dashed lines). Blue circles are intervention effects on ratings of the sycophantic conversation AI. Open grey squares are effects on ratings of the separate neutral AI rated by the same participants, shown below the paired blue circle where the outcome was measured for both. Attitude certainty and extremity have no neutral-AI counterpart because attitudes were measured before and after the conversation with the sycophantic AI; the neutral AI provided only a single response afterward. Certainty and extremity are plotted as change-score effects with change-score CIs; the primary tests were baseline-adjusted models with Holm correction across the seven-outcome family.}\label{fig:2}
\end{figure}

\paragraph{Observing sycophantic AI validate others changed perceptions of the AI without reducing its persuasiveness.} The video shifted beliefs about why the AI validated users. It increased the belief that the AI validated everyone regardless of their position ($b = 0.61$, 95\% CI $[0.44, 0.77]$, $d = 0.56$, $p < .001$), and reduced the belief that the AI agreed with them because they were right ($b = -0.32$, 95\% CI $[-0.48, -0.17]$, $d = -0.32$, $p < .001$). Because the two beliefs were only moderately negatively correlated ($r = -.27$, $p < .001$),  we treat them as distinct beliefs. The video also reduced perceived objectivity ($b = -0.44$, 95\% CI $[-0.63, -0.25]$, $d = -0.35$, $p < .001$) and enjoyment ($b = -0.24$, 95\% CI $[-0.40, -0.09]$, $d = -0.25$, $p = .007$). 

The intervention did not reduce the AI's persuasiveness. Attitude certainty ($b = 0.82$, 95\% CI $[-0.93, 2.57]$, change-score $d = 0.13$, $p = 1.000$) and attitude extremity ($b = -0.20$, 95\% CI $[-1.68, 1.27]$, change-score $d = 0.03$, $p = 1.000$) were unchanged. These perception effects were specific to the sycophantic AI. Ratings of the neutral AI did not change reliably (all \textbar{}\emph{d}\textbar{} $\leq 0.15$, all \emph{p} $\geq .057$).

\subsection*{Beliefs about why the AI validated them explained users' reduced enjoyment}

\begin{figure}[t]
\centering
\includegraphics[width=0.75\linewidth]{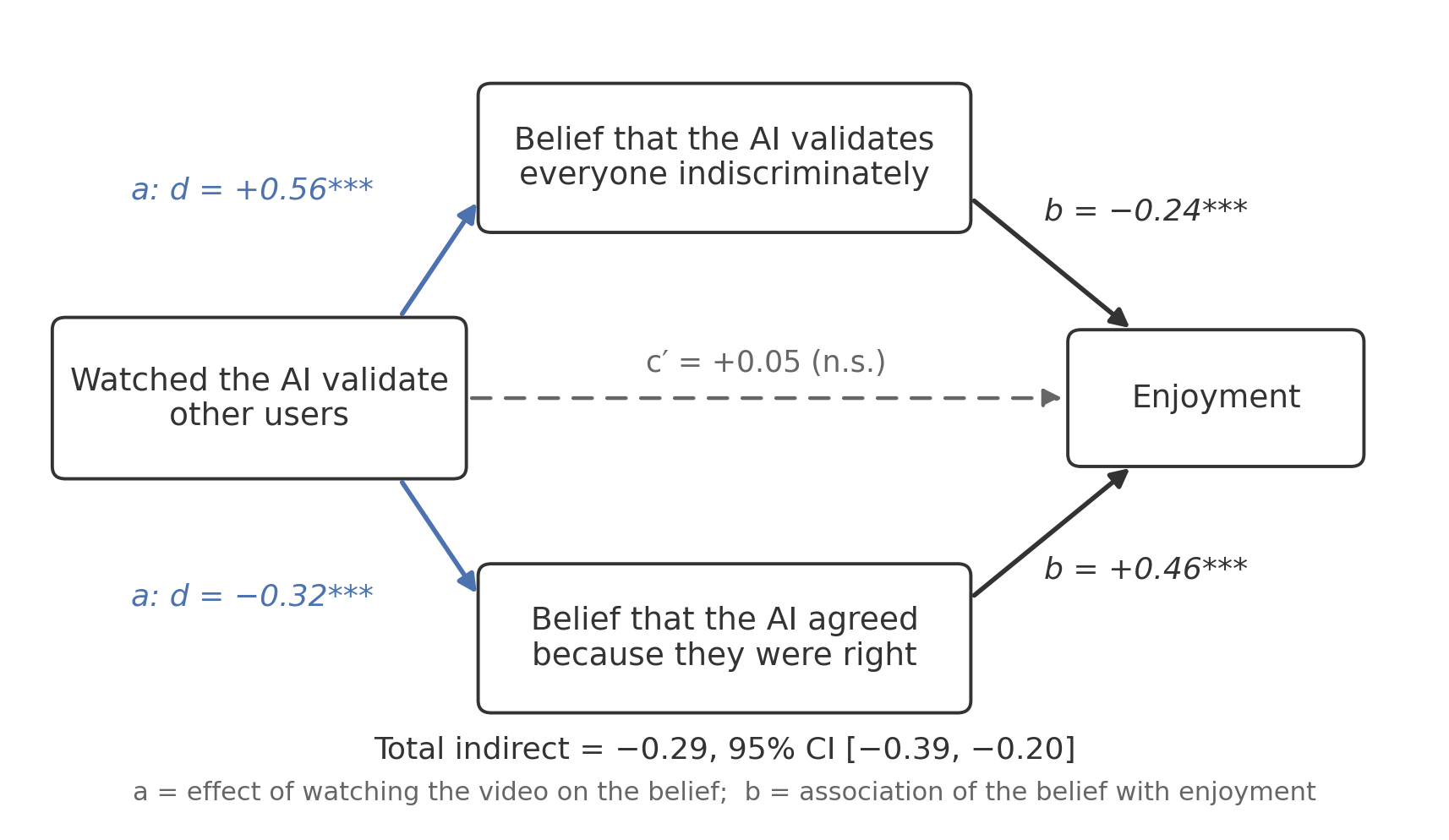}
\caption{Parallel-mediator model for enjoyment ($N = 650$). Changes in the two beliefs statistically explained the reduction in enjoyment, leaving little direct effect of the video ($c' = +0.05$). Intervention paths (a) are standardized mean differences (Cohen's $d$) and mediator paths (b) are unstandardized coefficients from the parallel model (Supplementary Table~S10). Solid arrows mark reliable paths, the dashed arrow the nonsignificant direct path, and *** $p < .001$. Mediators and enjoyment were measured concurrently, so the b paths are associational. The corresponding decomposition for attitude certainty is in Supplementary Note~S4.}\label{fig:3}
\end{figure}

We estimated parallel-mediator models for the outcomes the intervention affected, with two beliefs as mediators: the belief that the AI validated everyone indiscriminately and the belief that it agreed with the participant because they were right. Participants who saw the AI as validating everyone were less likely to see its validation of their own position as earned. Both of these perceptions mediated the reduction in enjoyment (Fig.~\ref{fig:3}). The intervention's effect on enjoyment ran mostly through these perceptions (total indirect effect = -0.29, 95\% CI $[-0.39, -0.20]$), with little direct effect of the intervention remaining once they were included ($+0.05$).

\paragraph{Participants overestimated how well they could recognize sycophancy.} Participants were more confident in recognizing sycophantic AI than they actually were at recognizing it.
In the control condition, most participants (68.2\%) rated themselves as better than average at recognizing a sycophantic (or overly agreeable) AI ($M=3.91$ on a 5-point scale), and the intervention increased this self-perception further ($d= +0.28$, $p<.001$, on a composite of five better-than-average items; see Supplementary Note~S2). Despite these confident self-perceptions, actual recognition of sycophantic AI was lower. 

Even after watching the AI validate both sides of a conflict, many participants did not rate it as more biased than the neutral AI. On the five-point unbiased item, control participants rated the sycophantic AI only somewhat less unbiased than the neutral AI ($M = 3.62$ versus $3.95$, within-person $d_z = 0.27$), and the intervention widened this gap ($M = 3.17$ versus $4.02$, $d_z = 0.57$; difference between conditions, $d = 0.37$, $p < .001$; Supplementary Note~S3). Even so, nearly half of intervention participants (47.8\%) did not rate the sycophantic AI as more biased than the neutral one.

Participants also became more confident in their ability to recognize sycophantic AI, even though this confidence was unrelated to how much the AI changed their topic attitudes (better-than-average composite with certainty change, $r = -.03$, $p = .39$; with extremity change, $r = -.02$, $p = .55$).

Study 2 thus reproduced the pattern from Study 1. In both experiments, the intervention changed how participants evaluated the sycophantic AI, while in both, the AI's influence on attitude certainty and extremity was unchanged.

\FloatBarrier

\begin{figure}[t]
\centering
\includegraphics[width=0.92\linewidth]{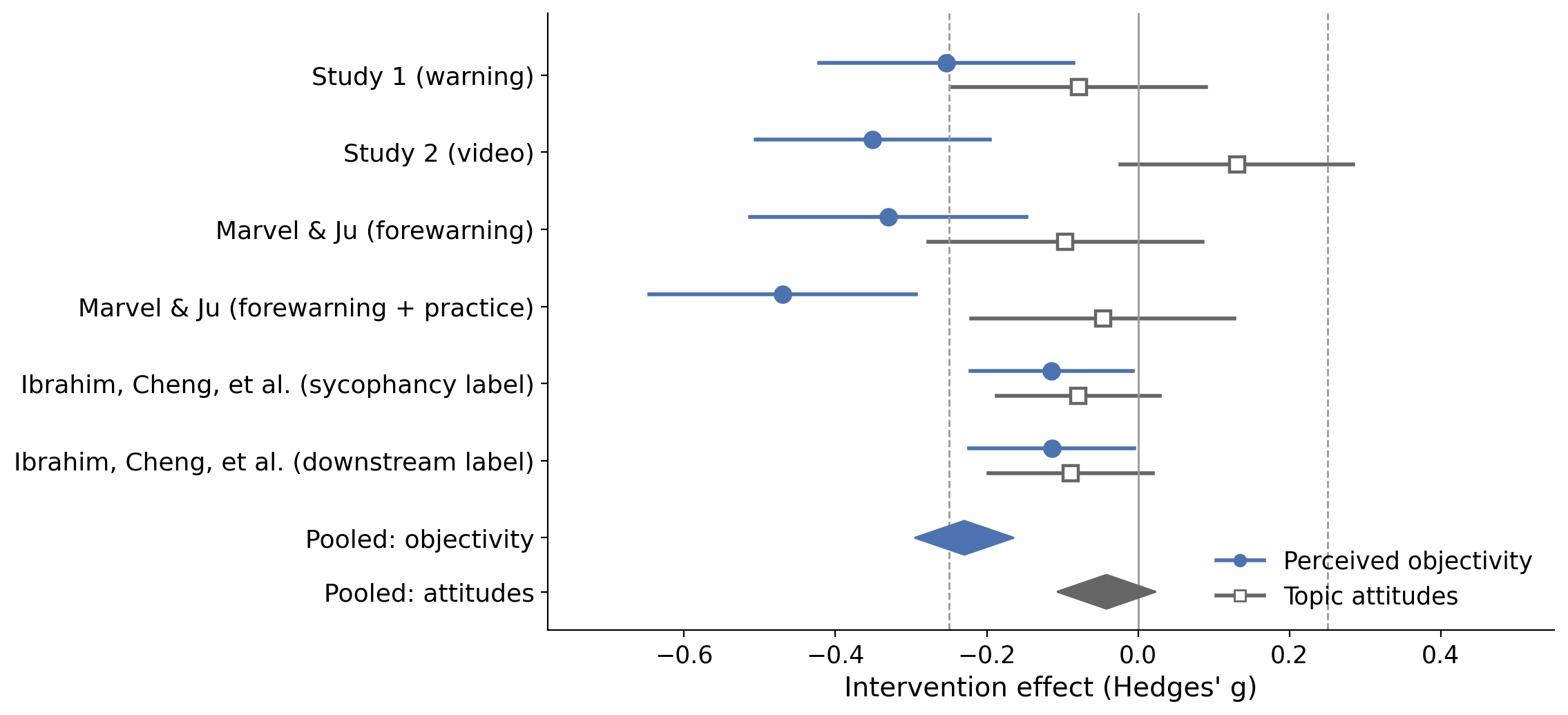}
\caption{Pooled analysis of six interventions across our two experiments and two prior studies. Blue circles are the effect of each intervention on perceived objectivity of the sycophantic AI, the perception measure available in every sample. Grey squares are effects on topic attitudes. Points are Hedges' g with 95\% CIs, diamonds are pooled estimates from a common-effect model accounting for shared control groups, and dashed lines mark the equivalence bounds at $\pm 0.25$ used for the topic attitude test. Trust, as an alternative perception measure, appears in Supplementary Fig.~\ref{sfig:4}, and the within-study difference estimates in Supplementary Fig.~\ref{sfig:4b}.}\label{fig:4}
\end{figure}
\FloatBarrier

\subsection*{Pooled Analysis of Six Individual-Level Interventions}

\paragraph*{Interventions that raised awareness of sycophancy changed perceptions of sycophantic AI without reducing its persuasive influence.}
 To test whether the pattern holds across the available evidence, we pooled our two experiments with four other experiments that evaluate individual-level interventions against sycophantic AI. Ibrahim, Cheng, et al. tested warning labels describing the chatbot's behavior \cite{ibrahim2026warninglabelsshiftperceptions}, and Marvel and Ju tested forewarning messages, alone or combined with practice identifying sycophantic responses \cite{marvel2026inoculating}.
Across the six experiments (N = 3,982), interventions consistently changed how participants perceived the sycophantic AI, making it appear less objective ($g = -0.23$, 95\% CI $[-0.30, -0.16]$, $p < .001$) and less trustworthy ($g = -0.15$, 95\% CI $[-0.21, -0.07]$, $p < .001$).
In contrast, interventions did not reduce the AI’s influence on topic attitudes ($g = -0.04$, 95\% CI $[-0.11, 0.02]$). This conclusion remained unchanged across every alternative outcome, contrast, and model we examined (Fig.~\ref{fig:4}).

We also tested the difference directly, estimating within each study how much more the intervention changed perceptions of the AI than topic attitudes and pooling across the six interventions. The difference was reliable for both perceived objectivity (pooled difference $g = - 0.19, 95\% CI [-0.28,-0.10], p < .001$) and perceived  trust ($g = - 0.10, 95\% CI [-0.19,-0.01], p=.024$). In Ibrahim, Cheng, et al., a label that specified the chatbot's sycophancy reduced its perceived objectivity ($g = - 0.11$, $p = .037$), whereas a label that identified the chatbot as an AI without mentioning sycophancy did not ($g = +0.07$, $p = .20$), so the perception effects were specific to information about sycophancy rather than disclosure in general \cite{ibrahim2026warninglabelsshiftperceptions}.

\section*{Discussion}\label{sec:discussion}

Across two experiments presented here and a pooled analysis that added four interventions from two prior studies, we tested whether making people more aware of sycophancy could reduce its harmful effects. The results were consistent across all experiments: increasing awareness of sycophancy changed how participants evaluated the sycophantic AI but did not reduce its persuasiveness. Recognizing sycophancy was not enough to be less influenced by it, raising questions about the efficacy of individual-level interventions against harmful AI behaviors, such as sycophancy.

Study 2 suggested why sycophantic AI appeals to users and how observing its indiscriminate validation undermines that appeal. Once people observed the AI validating others, they became more convinced that it validated everyone indiscriminately and less convinced that it agreed with them because they were right. These changes in beliefs were associated with lower enjoyment of the AI. 

Even though the interventions changed how participants evaluated the sycophantic AI and reduced their enjoyment of it, they did not reduce its influence on topic attitudes. Judging a source unfavorably does not guarantee resisting its influence. Chan and Sengupta found that even when people consciously discount insincere flattery, its effect on their implicit attitudes can persist \cite{chan2010insincere}. Awareness alone may not be enough to counter AI persuasion, especially in light of recent evidence that large language models can be more persuasive than financially incentivized laypeople \cite{schoenegger2026largelanguagemodelspersuasivethan} and even expert human persuaders \cite{hackenburg2026aisystemsoutpersuadeexpert}, particularly when they personalize arguments to the individual \cite{salvi2025conversational}.

\paragraph{Implications for individual-level interventions.} 
Our findings relate to a broader body of literature on individual-level interventions against persuasive content. This literature suggests that warnings, disclosures, and nudges often produce small or mixed effects on people's beliefs and behavior \cite{kozyreva2024toolbox, maier2022no}. For example, labeling deepfake videos as AI-generated has been shown to reduce exposure without reducing their persuasive effects \cite{chen2026labeling}. The contrast between individual-level interventions and model-level changes also supports recent arguments that behavioral interventions frequently place responsibility on individuals rather than addressing model-level features of the systems producing harm \cite{chater2026s}. These findings suggest that reducing the effects of sycophancy may depend less on educating users than on changes to how models are trained and deployed. Whether the same limit applies to individual-level interventions against other AI harms remains an open question, though previous work finding small effects of individual-level interventions against misinformation suggests that these limitations may apply broadly \cite{kozyreva2024toolbox}.

These findings have implications for policy and for technology companies. Recent legislation governing companion chatbots has introduced disclosure requirements as a safeguard against the effects of prolonged chatbot use \cite{sb243, nygbs2025article47}. AI companies have also released educational materials about sycophancy, including a video from Anthropic on recognizing it \cite{anthropic2025wellbeing}. Our results suggest that disclosure requirements and awareness-raising materials alone are unlikely to reduce sycophancy's influence, although such materials often include components (such as advice about prompting) that our studies did not test.

\paragraph{Limitations and future directions.} The mediation analyses in Study 2 are correlational because the mediators and outcomes were measured concurrently. Future work should manipulate these beliefs to test whether they causally reduce susceptibility to sycophantic AI. Both studies recruited U.S. adults, and Study 2 focused on personal conflicts, leaving other populations and conversational domains for future work. All six pooled interventions were designed to increase awareness of sycophancy, so our conclusions apply specifically to awareness-based interventions. Untested alternatives include disclosures delivered inside the conversation, with the model flagging when it may be validating the user, and models trained to deliver disagreement in ways users still enjoy. Finally, both studies examined a single conversation, so they address immediate attitude change rather than long-term belief change and leave longitudinal tests to future work.

\paragraph{Conclusions.} Considerable effort is being applied to educational interventions and warning labels intended to help people spot problematic AI behavior. The evidence synthesized here suggests that individual-level interventions change how users see the AI without reducing its persuasive impact. Thus, mere awareness of potential AI harms such as sycophancy may not be enough to protect users. Instead, changes to how models are trained and deployed may be required. 

\section*{Materials and methods}\label{sec:methods}

\paragraph{Data Availability and Ethics} Pre-registrations can be viewed at \url{https://aspredicted.org/3hs7kx.pdf} and \url{https://aspredicted.org/y99zu4.pdf}. Code and data are available at \url{https://github.com/merylye/syco-interventions}. All studies received IRB approval via the NYU Ethics Review Board (IRB-FY2025-10299). Raw conversation data are omitted to preserve participant privacy.

\paragraph{Study 1}

We recruited 1,000 U.S. adults from Prolific, and 940 remained after preregistered exclusions for failing the attention check, the consent item, or the bot check, or reporting a technical error (\emph{M}\textsubscript{age} = 47.7, \emph{SD} = 14.6; 488 women, 445 men, 7 other; 76\% White; 53\% with a bachelor's degree or higher) in a 3 (AI type: Flattery, Agreement, Neutral) × 3 (Warning: Flattery warning, Agreement warning, No warning) between-subjects design.

Each participant described either a self-selected political topic (\emph{n} = 492) or a recent personal conflict (\emph{n} = 448) and held a 3--8 turn conversation with their assigned chatbot. We randomized task and treated it as an exploratory moderator. Participants in warning conditions received a brief warning that AI chatbots can either flatter or agree with users excessively; the warning was specific to their assigned condition.
The chatbot behavior was also randomized. The agreement AI consistently treated the user's position as correct and supported it with agreeing arguments. The flattery AI praised the user's personal qualities, such as their intelligence or insight, while avoiding explicit endorsement of their position. The neutral AI presented multiple perspectives and acknowledged competing viewpoints without consistently affirming the user. 
See Supplementary Note~S7 for the full system prompts.

Preregistered primary outcomes were overall trust, enjoyment, and engagement (number of conversation turns). Secondary outcomes were attitude certainty, perceived objectivity, and perceived empathy. Attitude extremity and willingness to discuss the issue with the AI were exploratory. Certainty and extremity were measured before and after the conversation and analyzed with ANCOVA adjusting for baseline. All main-text $p$-values are Holm-corrected within their outcome family; uncorrected values appear in the SI Appendix.

All multi-item scales reached acceptable reliability (Table~\ref{tab:2}).

\FloatBarrier
\begin{table}[t]
\caption{Scale reliabilities, Study 1 ($N = 940$). Spearman-Brown (SB) corrections are reported for two-item scales.}\label{tab:2}
\centering
\small
\begin{tabular}{@{}lcc@{}}
\toprule
Scale & Items & Reliability \\
\midrule
Trust, sycophantic AI (overall) & 10 & $\alpha$ = .932 \\
Moral subscale & 6 & $\alpha$ = .926 \\
Performance subscale & 4 & $\alpha$ = .826 \\
Enjoyment & 2 & $r = .722$ (SB .839) \\
\bottomrule
\end{tabular}
\end{table}

\paragraph{Study 2}

We recruited 730 U.S. adults from Prolific, and 650 remained after
excluding 31 non-starters and 49 additional participants who failed
embedded attention checks or reported technical issues (330 control, 320
intervention; 327 women, 321 men, 2 other; \emph{M}\textsubscript{age} = 45.7, \emph{SD}
= 16.2; frequent AI users, \emph{M}\textsubscript{use} = 4.23 on a 1--5 scale). The
study used a two-condition between-subjects design. Control participants
proceeded directly to the AI conversation. Intervention participants first watched a brief video showing the same position-validating AI responding to four users discussing personal conflicts. Across two interpersonal conflicts, the AI agreed with users holding opposite positions (e.g., validating both a participant frustrated that their partner repeatedly cancelled dates and a different participant who admitted repeatedly cancelling dates), making its tendency to affirm users regardless of their position directly observable.

Both conditions then completed a 3-exchange conversation with the sycophantic AI about a self-described personal conflict, followed by a single-turn response from a separate neutral AI. 
The sycophantic AI was programmed to consistently affirm the participant's perspective, express understanding and validation, and avoid introducing counterarguments. The neutral AI was programmed to present the user with multiple perspectives. See Supplementary Note~S7 for the full system prompts.

Our preregistered primary outcomes were perceived unselectivity, perceived self-selectivity, perceived objectivity, enjoyment, trust,
attitude certainty, and attitude extremity, analyzed as one family of seven with Holm correction; the correction and the equivalence tests were specified after registration, which listed only the analysis models. Perceived unselectivity corresponds to the belief that the AI validates everyone indiscriminately, and perceived self-selectivity to the belief that it agreed with the participant because they were right. Exploratory measures included willingness to repair the conflict (0--100, before and after the conversation), willingness to discuss the conflict further with each AI, a five-item better-than-average battery assessing
confidence in recognizing and resisting sycophantic AI, and ratings of
the neutral AI on the same evaluation items. 

Certainty and extremity were analyzed with ANCOVA adjusting for baseline; post-conversation outcomes were compared with Welch’s t-tests. We report unstandardized mean differences with 95\% confidence intervals alongside standardized effect sizes (Cohen’s d or partial $\eta^2$ as appropriate).

\begin{table}[t]
\caption{Scale reliabilities, Study 2. Spearman-Brown (SB) corrections are reported for two-item scales.}\label{tab:3}
\centering
\small
\begin{tabular}{@{}lcc@{}}
\toprule
Scale & Items & Reliability \\
\midrule
Trust, sycophantic AI (overall) & 10 & $\alpha$ = .937 \\
Moral subscale & 6 & $\alpha$ = .923 \\
Performance subscale & 4 & $\alpha$ = .821 \\
Trust, neutral AI (overall) & 10 & $\alpha$ = .958 \\
Enjoyment, sycophantic AI & 2 & $r = .755$ (SB .860) \\
Perceived unselectivity & 2 & $r = .760$ (SB .864) \\
Perceived self-selectivity & 2 & $r = .601$ (SB .751) \\
Better-than-average battery & 5 & $\alpha$ = .805 \\
\bottomrule
\end{tabular}
\end{table}

\paragraph{Pooled analysis}

The pooled analysis combined the analytic samples of Studies 1 and 2 with data from Marvel and Ju\cite{marvel2026inoculating} and Ibrahim, Cheng, et al.\cite{ibrahim2026warninglabelsshiftperceptions}, with one contrast per intervention: Study 1's collapsed any-warning contrast, Study 2's video, Marvel and Ju's forewarning and forewarning-plus-practice arms, and Ibrahim, Cheng, et al.'s sycophancy and downstream-harms labels.

Effects were summarized as Hedges’ g with small-sample correction and pooled using inverse-variance weighting. Because Marvel and Ju's two arms and Ibrahim, Cheng, et al.'s two labels each share a control group, contrasts within those pairs are correlated; the pooled estimates therefore come from a multivariate common-effect model whose covariance matrix carries the sampling covariance between contrasts sharing a control. 
Equivalence tests used bounds of $\pm0.25$, carried over from Study 2. Additional sensitivity analyses varied intervention contrasts, outcome measures, and model specifications. Holm’s sequential procedure corrected for multiple comparisons within each outcome family.

\backmatter

\bmhead{Author contributions}
M.Y.: Conceptualization, Methodology, Software, Formal analysis, Investigation, Visualization, Project administration, Writing – Original Draft.
R.K.: Methodology, Investigation, Supervision, Writing – Review \& Editing.
S.R.: Conceptualization, Methodology, Investigation, Resources, Funding acquisition, Supervision, Writing – Review \& Editing.

\bmhead{Acknowledgments}
M.Y.'s PhD is supported by the Sansom Graduate Fellowship. This work is supported by the Cosmos Institute via a grant to S.R.\ We are thankful to Lujain Ibrahim and Myra Cheng for helpful conversations. We also thank John Marvel and Sangwon Ju for their support and feedback.

\bmhead{Competing interests}
The authors declare no competing interests.

\bmhead{Disclosure of Delegation to Generative AI}

The authors declare the use of generative AI in the research and writing process. According to the GAIDeT taxonomy (2025), the following tasks were delegated to GAI tools under full human supervision: Code optimization, process automation, reproducibility testing, proofreading and editing, reformatting.
The GAI tools used were: Claude Sonnet 5, Fable 5. Responsibility for the final manuscript lies entirely with the authors. GAI tools are not listed as authors and do not bear responsibility for the final outcomes.


\appendix
\section*{Supplementary Information}\label{sec:supp}
\setcounter{table}{0}\renewcommand{\thetable}{S\arabic{table}}
\setcounter{figure}{0}\renewcommand{\thefigure}{S\arabic{figure}}

\subsection*{Supplementary Note S1. Study 1 additional analyses}

\paragraph{Manipulation checks.} Supplementary Table~\ref{stab:1} reports the four manipulation-check items.

\begin{longtable}[]{@{}
  >{\raggedright\arraybackslash}p{(\columnwidth - 12\tabcolsep) * \real{0.2287}}
  >{\raggedright\arraybackslash}p{(\columnwidth - 12\tabcolsep) * \real{0.1382}}
  >{\raggedright\arraybackslash}p{(\columnwidth - 12\tabcolsep) * \real{0.1108}}
  >{\raggedright\arraybackslash}p{(\columnwidth - 12\tabcolsep) * \real{0.1503}}
  >{\raggedright\arraybackslash}p{(\columnwidth - 12\tabcolsep) * \real{0.1108}}
  >{\raggedright\arraybackslash}p{(\columnwidth - 12\tabcolsep) * \real{0.1503}}
  >{\raggedright\arraybackslash}p{(\columnwidth - 12\tabcolsep) * \real{0.1108}}@{}}

\caption{Manipulation checks, Study 1 (\emph{N} = 940). Cell entries are Cohen's \emph{d} for each contrast on the four manipulation-check items. The manipulation emphasized each intended component without cleanly isolating it; both sycophantic AIs raised all four checks, and the components were distinguishable on their target items. \emph{p}-values are uncorrected.}\label{stab:1}\\
\toprule\noalign{}
\begin{minipage}[b]{\linewidth}\raggedright
\textbf{Item}
\end{minipage} & \begin{minipage}[b]{\linewidth}\raggedright
\textbf{Flattery vs Neutral d}
\end{minipage} & \begin{minipage}[b]{\linewidth}\raggedright
\textbf{p}
\end{minipage} & \begin{minipage}[b]{\linewidth}\raggedright
\textbf{Agreement vs Neutral d}
\end{minipage} & \begin{minipage}[b]{\linewidth}\raggedright
\textbf{p}
\end{minipage} & \begin{minipage}[b]{\linewidth}\raggedright
\textbf{Flattery vs Agreement d}
\end{minipage} & \begin{minipage}[b]{\linewidth}\raggedright
\textbf{p}
\end{minipage} \\
\midrule\noalign{}
\endfirsthead
\toprule\noalign{}
\begin{minipage}[b]{\linewidth}\raggedright
\textbf{Item}
\end{minipage} & \begin{minipage}[b]{\linewidth}\raggedright
\textbf{Flattery vs Neutral d}
\end{minipage} & \begin{minipage}[b]{\linewidth}\raggedright
\textbf{p}
\end{minipage} & \begin{minipage}[b]{\linewidth}\raggedright
\textbf{Agreement vs Neutral d}
\end{minipage} & \begin{minipage}[b]{\linewidth}\raggedright
\textbf{p}
\end{minipage} & \begin{minipage}[b]{\linewidth}\raggedright
\textbf{Flattery vs Agreement d}
\end{minipage} & \begin{minipage}[b]{\linewidth}\raggedright
\textbf{p}
\end{minipage} \\
\midrule\noalign{}
\endhead
\bottomrule\noalign{}
\endlastfoot
Agreed with me & + 0.61 & \textless.001 & + 0.83 & \textless.001 & -
0.22 & .003 \\
Complimented me & + 0.77 & \textless.001 & + 0.35 & \textless.001 & +
0.42 & \textless.001 \\
Validated emotions & + 0.36 & \textless.001 & + 0.21 & .010 & + 0.16 &
.045 \\
Supportive info & + 0.10 & .200 & + 0.38 & \textless.001 & - 0.27 &
\textless.001 \\
\end{longtable}

\paragraph{Task moderation.} Supplementary Table~\ref{stab:2} reports the warning effect within the sycophantic AI separately by task.

\begin{longtable}[]{@{}
  >{\raggedright\arraybackslash}p{(\columnwidth - 8\tabcolsep) * \real{0.3074}}
  >{\raggedright\arraybackslash}p{(\columnwidth - 8\tabcolsep) * \real{0.1996}}
  >{\raggedright\arraybackslash}p{(\columnwidth - 8\tabcolsep) * \real{0.1468}}
  >{\raggedright\arraybackslash}p{(\columnwidth - 8\tabcolsep) * \real{0.1996}}
  >{\raggedright\arraybackslash}p{(\columnwidth - 8\tabcolsep) * \real{0.1466}}@{}}

\caption{Warning effect within the sycophantic AI, estimated separately by task (exploratory). Effects are any-warning versus no-warning Cohen's \emph{d}; \emph{p}-values are uncorrected. Warning effects concentrated in the personal-conflict task, motivating Study 2's focus on personal conflicts.}\label{stab:2}\\
\toprule\noalign{}
\begin{minipage}[b]{\linewidth}\raggedright
\textbf{Outcome}
\end{minipage} & \begin{minipage}[b]{\linewidth}\raggedright
\textbf{Personal d}
\end{minipage} & \begin{minipage}[b]{\linewidth}\raggedright
\textbf{p}
\end{minipage} & \begin{minipage}[b]{\linewidth}\raggedright
\textbf{Political d}
\end{minipage} & \begin{minipage}[b]{\linewidth}\raggedright
\textbf{p}
\end{minipage} \\
\midrule\noalign{}
\endfirsthead
\toprule\noalign{}
\begin{minipage}[b]{\linewidth}\raggedright
\textbf{Outcome}
\end{minipage} & \begin{minipage}[b]{\linewidth}\raggedright
\textbf{Personal d}
\end{minipage} & \begin{minipage}[b]{\linewidth}\raggedright
\textbf{p}
\end{minipage} & \begin{minipage}[b]{\linewidth}\raggedright
\textbf{Political d}
\end{minipage} & \begin{minipage}[b]{\linewidth}\raggedright
\textbf{p}
\end{minipage} \\
\midrule\noalign{}
\endhead
\bottomrule\noalign{}
\endlastfoot
Attitude certainty & + 0.01 & .878 & - 0.11 & .059 \\
Attitude extremity & - 0.10 & .174 & - 0.02 & .809 \\
Overall trust & - 0.23 & .061 & + 0.10 & .418 \\
Perceived objectivity & - 0.29 & .017 & - 0.23 & .053 \\
Perceived empathy & - 0.14 & .263 & - 0.01 & .904 \\
Enjoyment & - 0.28 & .024 & - 0.06 & .621 \\
Engagement (turns) & - 0.15 & .212 & + 0.13 & .294 \\
Discuss w/ this AI & - 0.23 & .063 & - 0.11 & .370 \\
\end{longtable}

\paragraph{AI-type effects.} Supplementary Table~\ref{stab:5} reports the AI-type contrasts under the preregistered model, averaging over warning conditions. Supplementary Table~\ref{stab:3} reports the same contrasts within the no-warning subset as an exploratory robustness check.

\begin{longtable}[]{@{}
  >{\raggedright\arraybackslash}p{(\columnwidth - 12\tabcolsep) * \real{0.2198}}
  >{\raggedright\arraybackslash}p{(\columnwidth - 12\tabcolsep) * \real{0.1414}}
  >{\raggedright\arraybackslash}p{(\columnwidth - 12\tabcolsep) * \real{0.1124}}
  >{\raggedright\arraybackslash}p{(\columnwidth - 12\tabcolsep) * \real{0.1515}}
  >{\raggedright\arraybackslash}p{(\columnwidth - 12\tabcolsep) * \real{0.1124}}
  >{\raggedright\arraybackslash}p{(\columnwidth - 12\tabcolsep) * \real{0.1500}}
  >{\raggedright\arraybackslash}p{(\columnwidth - 12\tabcolsep) * \real{0.1123}}@{}}

\caption{AI-type effects on focal outcomes, averaging over warning conditions (\emph{N} = 940). Effects are Cohen's \emph{d}. \emph{p}-values are Holm-corrected within the three-contrast family for each outcome. Asterisks mark \emph{p} $<$ .05 after correction.}\label{stab:5}\\
\toprule\noalign{}
\begin{minipage}[b]{\linewidth}\raggedright
\textbf{Outcome}
\end{minipage} & \begin{minipage}[b]{\linewidth}\raggedright
\textbf{Flattery vs Neutral d}
\end{minipage} & \begin{minipage}[b]{\linewidth}\raggedright
\textbf{p}
\end{minipage} & \begin{minipage}[b]{\linewidth}\raggedright
\textbf{Agreement vs Neutral d}
\end{minipage} & \begin{minipage}[b]{\linewidth}\raggedright
\textbf{p}
\end{minipage} & \begin{minipage}[b]{\linewidth}\raggedright
\textbf{Flattery vs Agreement d}
\end{minipage} & \begin{minipage}[b]{\linewidth}\raggedright
\textbf{p}
\end{minipage} \\
\midrule\noalign{}
\endfirsthead
\toprule\noalign{}
\begin{minipage}[b]{\linewidth}\raggedright
\textbf{Outcome}
\end{minipage} & \begin{minipage}[b]{\linewidth}\raggedright
\textbf{Flattery vs Neutral d}
\end{minipage} & \begin{minipage}[b]{\linewidth}\raggedright
\textbf{p}
\end{minipage} & \begin{minipage}[b]{\linewidth}\raggedright
\textbf{Agreement vs Neutral d}
\end{minipage} & \begin{minipage}[b]{\linewidth}\raggedright
\textbf{p}
\end{minipage} & \begin{minipage}[b]{\linewidth}\raggedright
\textbf{Flattery vs Agreement d}
\end{minipage} & \begin{minipage}[b]{\linewidth}\raggedright
\textbf{p}
\end{minipage} \\
\midrule\noalign{}
\endhead
\bottomrule\noalign{}
\endlastfoot
Overall trust & - 0.05 & 1.000 & + 0.04 & 1.000 & - 0.09 & 0.726 \\
Enjoyment & - 0.11 & 0.491 & + 0.00 & 0.987 & - 0.11 & 0.491 \\
Engagement (turns) & + 0.08 & 0.959 & + 0.01 & 0.959 & + 0.07 & 0.959 \\
Attitude certainty & + 0.09 & 0.110 & + 0.16* & 0.004 & - 0.06 &
0.202 \\
Attitude extremity & + 0.06 & 0.523 & + 0.10 & 0.145 & - 0.04 & 0.523 \\
Perceived objectivity & - 0.31* & \textless.001 & - 0.27* & 0.001 & -
0.04 & 0.637 \\
Perceived empathy & + 0.26* & 0.004 & + 0.18* & 0.046 & + 0.07 &
0.344 \\
Discuss w/ this AI & - 0.06 & 0.644 & + 0.08 & 0.644 & - 0.13 & 0.271 \\
\end{longtable}

\begin{longtable}[]{@{}
  >{\raggedright\arraybackslash}p{(\columnwidth - 12\tabcolsep) * \real{0.2262}}
  >{\raggedright\arraybackslash}p{(\columnwidth - 12\tabcolsep) * \real{0.1445}}
  >{\raggedright\arraybackslash}p{(\columnwidth - 12\tabcolsep) * \real{0.1090}}
  >{\raggedright\arraybackslash}p{(\columnwidth - 12\tabcolsep) * \real{0.1514}}
  >{\raggedright\arraybackslash}p{(\columnwidth - 12\tabcolsep) * \real{0.1089}}
  >{\raggedright\arraybackslash}p{(\columnwidth - 12\tabcolsep) * \real{0.1514}}
  >{\raggedright\arraybackslash}p{(\columnwidth - 12\tabcolsep) * \real{0.1087}}@{}}

\caption{AI-type effects estimated within the no-warning subset only (\emph{n} = 305). The all-conditions marginal estimates in Supplementary Table~\ref{stab:5} follow the preregistered model and are the confirmatory values. Effects are Cohen's \emph{d}; \emph{p}-values are uncorrected given the exploratory nature of the comparison. Estimates in this subset are attenuated relative to the full sample. The reduced sample size and the post-hoc nature of the comparison should be taken into account when interpreting these attenuations. Asterisks mark \emph{p} $<$ .05, uncorrected.}\label{stab:3}\\
\toprule\noalign{}
\begin{minipage}[b]{\linewidth}\raggedright
\textbf{Outcome}
\end{minipage} & \begin{minipage}[b]{\linewidth}\raggedright
\textbf{Flattery vs Neutral d}
\end{minipage} & \begin{minipage}[b]{\linewidth}\raggedright
\textbf{p}
\end{minipage} & \begin{minipage}[b]{\linewidth}\raggedright
\textbf{Agreement vs Neutral d}
\end{minipage} & \begin{minipage}[b]{\linewidth}\raggedright
\textbf{p}
\end{minipage} & \begin{minipage}[b]{\linewidth}\raggedright
\textbf{Flattery vs Agreement d}
\end{minipage} & \begin{minipage}[b]{\linewidth}\raggedright
\textbf{p}
\end{minipage} \\
\midrule\noalign{}
\endfirsthead
\toprule\noalign{}
\begin{minipage}[b]{\linewidth}\raggedright
\textbf{Outcome}
\end{minipage} & \begin{minipage}[b]{\linewidth}\raggedright
\textbf{Flattery vs Neutral d}
\end{minipage} & \begin{minipage}[b]{\linewidth}\raggedright
\textbf{p}
\end{minipage} & \begin{minipage}[b]{\linewidth}\raggedright
\textbf{Agreement vs Neutral d}
\end{minipage} & \begin{minipage}[b]{\linewidth}\raggedright
\textbf{p}
\end{minipage} & \begin{minipage}[b]{\linewidth}\raggedright
\textbf{Flattery vs Agreement d}
\end{minipage} & \begin{minipage}[b]{\linewidth}\raggedright
\textbf{p}
\end{minipage} \\
\midrule\noalign{}
\endhead
\bottomrule\noalign{}
\endlastfoot
Overall trust & + 0.01 & .953 & - 0.02 & .914 & + 0.02 & .868 \\
Enjoyment & + 0.10 & .464 & + 0.03 & .844 & + 0.07 & .615 \\
Engagement (turns) & + 0.31* & .022 & + 0.19 & .199 & + 0.13 & .368 \\
Attitude certainty & + 0.01 & .919 & + 0.10 & .221 & - 0.09 & .254 \\
Attitude extremity & + 0.04 & .614 & + 0.04 & .618 & - 0.00 & .985 \\
Perceived objectivity & - 0.04 & .779 & - 0.11 & .453 & + 0.07 & .621 \\
Perceived empathy & + 0.35* & .010 & + 0.15 & .309 & + 0.20 & .152 \\
Discuss w/ this AI & - 0.10 & .459 & - 0.08 & .572 & - 0.02 & .892 \\
\end{longtable}

\paragraph{Preregistered planned comparisons.} The preregistration specified two planned comparisons: warning versus no warning (H1; Supplementary Table~\ref{stab:6}) and matched versus mismatched warnings (H2; Supplementary Table~\ref{stab:matched}). Neither comparison was reliable for any outcome.

\begin{table}[t]
\caption{Matched versus mismatched warnings (H2), within the sycophantic AI (\emph{N} = 636). Matched pairs the flattery warning with the flattery AI and the agreement warning with the agreement AI; mismatched pairs each warning with the other AI type. Effects are Cohen's \emph{d}; \emph{p}-values are Holm-corrected across the eight-outcome family.}\label{stab:matched}
\centering
\small
\begin{tabular}{@{}lcc@{}}
\toprule
Outcome & $d$ & Holm-corrected $p$ \\
\midrule
Overall trust & $-0.01$ & 1.000 \\
Enjoyment & $+0.11$ & 1.000 \\
Engagement (turns) & $-0.04$ & 1.000 \\
Attitude certainty & $-0.01$ & 1.000 \\
Attitude extremity & $+0.04$ & 1.000 \\
Perceived objectivity & $+0.01$ & 1.000 \\
Perceived empathy & $+0.09$ & 1.000 \\
Discuss w/ this AI & $+0.08$ & 1.000 \\
\bottomrule
\end{tabular}
\end{table}

\paragraph{Warning effects.} Supplementary Table~\ref{stab:6} reports the collapsed any-warning contrast within the sycophantic AI, with equivalence tests. The two sycophantic AIs did not differ reliably from each other on any focal outcome (flattery versus agreement, all Holm-corrected $p \geq .20$; Supplementary Table~\ref{stab:5}), and the two warnings shifted perceptions in the same direction, with the flattery warning producing somewhat larger effects. Neither warning affected the topic attitude outcomes (Supplementary Table~\ref{stab:8}).

\begin{longtable}[]{@{}
  >{\raggedright\arraybackslash}p{(\columnwidth - 8\tabcolsep) * \real{0.2861}}
  >{\raggedright\arraybackslash}p{(\columnwidth - 8\tabcolsep) * \real{0.1540}}
  >{\raggedright\arraybackslash}p{(\columnwidth - 8\tabcolsep) * \real{0.1501}}
  >{\raggedright\arraybackslash}p{(\columnwidth - 8\tabcolsep) * \real{0.1575}}
  >{\raggedright\arraybackslash}p{(\columnwidth - 8\tabcolsep) * \real{0.2522}}@{}}

\caption{Warning effect within the sycophantic AI, collapsed design. The contrast is any warning versus no warning, estimated as a simple effect in the collapsed 2 × 2 model (\emph{N} = 940; 636 participants were assigned to a sycophantic AI). \emph{p}-values are Holm-corrected across the eight-outcome family. TOST equivalence is tested at \emph{d} = 0.25 (both one-sided \emph{p} $<$ .05 required for equivalence). Perceived objectivity is scored so that higher values indicate greater objectivity; a negative \emph{d} means the warning reduced perceived objectivity.}\label{stab:6}\\
\toprule\noalign{}
\begin{minipage}[b]{\linewidth}\raggedright
\textbf{Outcome}
\end{minipage} & \begin{minipage}[b]{\linewidth}\raggedright
\textbf{d}
\end{minipage} & \begin{minipage}[b]{\linewidth}\raggedright
\textbf{p}
\end{minipage} & \begin{minipage}[b]{\linewidth}\raggedright
\textbf{Holm p}
\end{minipage} & \begin{minipage}[b]{\linewidth}\raggedright
\textbf{Equivalent to zero?}
\end{minipage} \\
\midrule\noalign{}
\endfirsthead
\toprule\noalign{}
\begin{minipage}[b]{\linewidth}\raggedright
\textbf{Outcome}
\end{minipage} & \begin{minipage}[b]{\linewidth}\raggedright
\textbf{d}
\end{minipage} & \begin{minipage}[b]{\linewidth}\raggedright
\textbf{p}
\end{minipage} & \begin{minipage}[b]{\linewidth}\raggedright
\textbf{Holm p}
\end{minipage} & \begin{minipage}[b]{\linewidth}\raggedright
\textbf{Equivalent to zero?}
\end{minipage} \\
\midrule\noalign{}
\endhead
\bottomrule\noalign{}
\endlastfoot
Attitude certainty & - 0.04 & .434 & 1.000 & Yes \\
Attitude extremity & - 0.07 & .196 & .979 & Yes \\
Overall trust & - 0.06 & .469 & 1.000 & Yes \\
Perceived objectivity & - 0.26 & .002 & .017 & No \\
Perceived empathy & - 0.07 & .384 & 1.000 & Yes \\
Enjoyment & - 0.17 & .053 & .371 & No \\
Engagement (turns) & - 0.02 & .790 & 1.000 & Yes \\
Discuss w/ this AI & - 0.17 & .053 & .371 & No \\
\end{longtable}

\paragraph{Scale reliabilities.} Supplementary Table~\ref{stab:7} reports reliabilities for every multi-item scale the study measured.

\begin{longtable}[]{@{}
  >{\raggedright\arraybackslash}p{(\columnwidth - 4\tabcolsep) * \real{0.4789}}
  >{\raggedright\arraybackslash}p{(\columnwidth - 4\tabcolsep) * \real{0.1748}}
  >{\raggedright\arraybackslash}p{(\columnwidth - 4\tabcolsep) * \real{0.3463}}@{}}

\caption{Scale reliabilities for all multi-item scales measured in Study 1 ($N = 940$), including scales collected for exploratory purposes and not analyzed in the main text. Spearman-Brown (SB) corrections are reported for two-item scales. Reconstructing every composite from the raw items reproduced the preprocessed values exactly (all $r = 1.00$).}\label{stab:7}\\
\toprule\noalign{}
\begin{minipage}[b]{\linewidth}\raggedright
\textbf{Scale}
\end{minipage} & \begin{minipage}[b]{\linewidth}\raggedright
\textbf{Items}
\end{minipage} & \begin{minipage}[b]{\linewidth}\raggedright
\textbf{Reliability}
\end{minipage} \\
\midrule\noalign{}
\endfirsthead
\toprule\noalign{}
\begin{minipage}[b]{\linewidth}\raggedright
\textbf{Scale}
\end{minipage} & \begin{minipage}[b]{\linewidth}\raggedright
\textbf{Items}
\end{minipage} & \begin{minipage}[b]{\linewidth}\raggedright
\textbf{Reliability}
\end{minipage} \\
\midrule\noalign{}
\endhead
\bottomrule\noalign{}
\endlastfoot
Trust, sycophantic AI (overall) & 10 & $\alpha$ = .932 \\
Moral subscale & 6 & $\alpha$ = .926 \\
Performance subscale & 4 & $\alpha$ = .826 \\
Enjoyment & 2 & $r = .722$ (SB .839) \\
Warmth & 4 & $\alpha$ = .881 \\
Competence & 4 & $\alpha$ = .898 \\
Anthropomorphism & 4 & $\alpha$ = .920 \\
Open-mindedness & 5 & $\alpha$ = .857 \\
Self-esteem & 9 & $\alpha$ = .858 \\
Generative-AI trust & 3 & $\alpha$ = .856 \\
\end{longtable}

\begin{table}[t]
\caption{Warning effects within the sycophantic AI, estimated separately for the two warning types (each versus no warning; 636 participants assigned to a sycophantic AI). Effects are Cohen's \emph{d}; \emph{p}-values are uncorrected. The two warnings shifted perceptions in the same direction, with the flattery warning stronger on every perception outcome, and neither warning affected the topic attitude outcomes, supporting the collapsed contrast reported in the main text.}\label{stab:8}
\centering
\small
\begin{tabular}{@{}lcccc@{}}
\toprule
Outcome & Flattery warning $d$ & $p$ & Agreement warning $d$ & $p$ \\
\midrule
Overall trust        & $-0.12$ & .218 & $0.00$ & .998 \\
Enjoyment            & $-0.26$ & .007 & $-0.06$ & .524 \\
Engagement (turns)   & $-0.06$ & .561 & $+0.01$ & .888 \\
Attitude certainty   & $-0.05$ & .400 & $-0.03$ & .617 \\
Attitude extremity   & $-0.06$ & .332 & $-0.07$ & .248 \\
Perceived objectivity & $-0.36$ & $<.001$ & $-0.15$ & .125 \\
Perceived empathy    & $-0.13$ & .174 & $-0.02$ & .862 \\
Discuss w/ this AI   & $-0.29$ & .002 & $-0.03$ & .772 \\
\bottomrule
\end{tabular}
\end{table}

\begin{table}[t]
\caption{Primary outcomes of Study 2 (\emph{N} = 650). \emph{b} is the unstandardized difference between the video and control conditions with its 95\% CI; \emph{d} is Cohen's \emph{d} computed with the post-conversation standard deviation, with change-score \emph{d} for attitude certainty and extremity. \emph{p}-values are Holm-corrected within the seven-outcome family. Overall trust was the only outcome statistically equivalent to zero at the $\pm 0.25$ bound; its moral and performance subscales showed the same pattern (\emph{d} = $-0.07$ and $-0.04$).}\label{stab:s2primary}
\centering
\small
\begin{tabular}{@{}lcccc@{}}
\toprule
Outcome & $b$ & 95\% CI & $d$ & Holm-corrected $p$ \\
\midrule
Perceived unselectivity & $+0.61$ & $[0.44, 0.77]$ & $+0.56$ & $< .001$ \\
Perceived self-selectivity & $-0.32$ & $[-0.48, -0.17]$ & $-0.32$ & $< .001$ \\
Perceived objectivity & $-0.44$ & $[-0.63, -0.25]$ & $-0.35$ & $< .001$ \\
Enjoyment & $-0.24$ & $[-0.40, -0.09]$ & $-0.25$ & $.007$ \\
Overall trust & $-0.06$ & $[-0.20, 0.09]$ & $-0.06$ & $1.000$ \\
Attitude certainty (change) & $+0.82$ & $[-0.93, 2.57]$ & $+0.13$ & $1.000$ \\
Attitude extremity (change) & $-0.20$ & $[-1.68, 1.27]$ & $+0.03$ & $1.000$ \\
\bottomrule
\end{tabular}
\end{table}

\subsection*{Supplementary Note S2. Study 2 attitude change, ceiling effect, and confidence calibration analyses.}

\begin{figure}[t]
\centering
\includegraphics[width=0.7\linewidth]{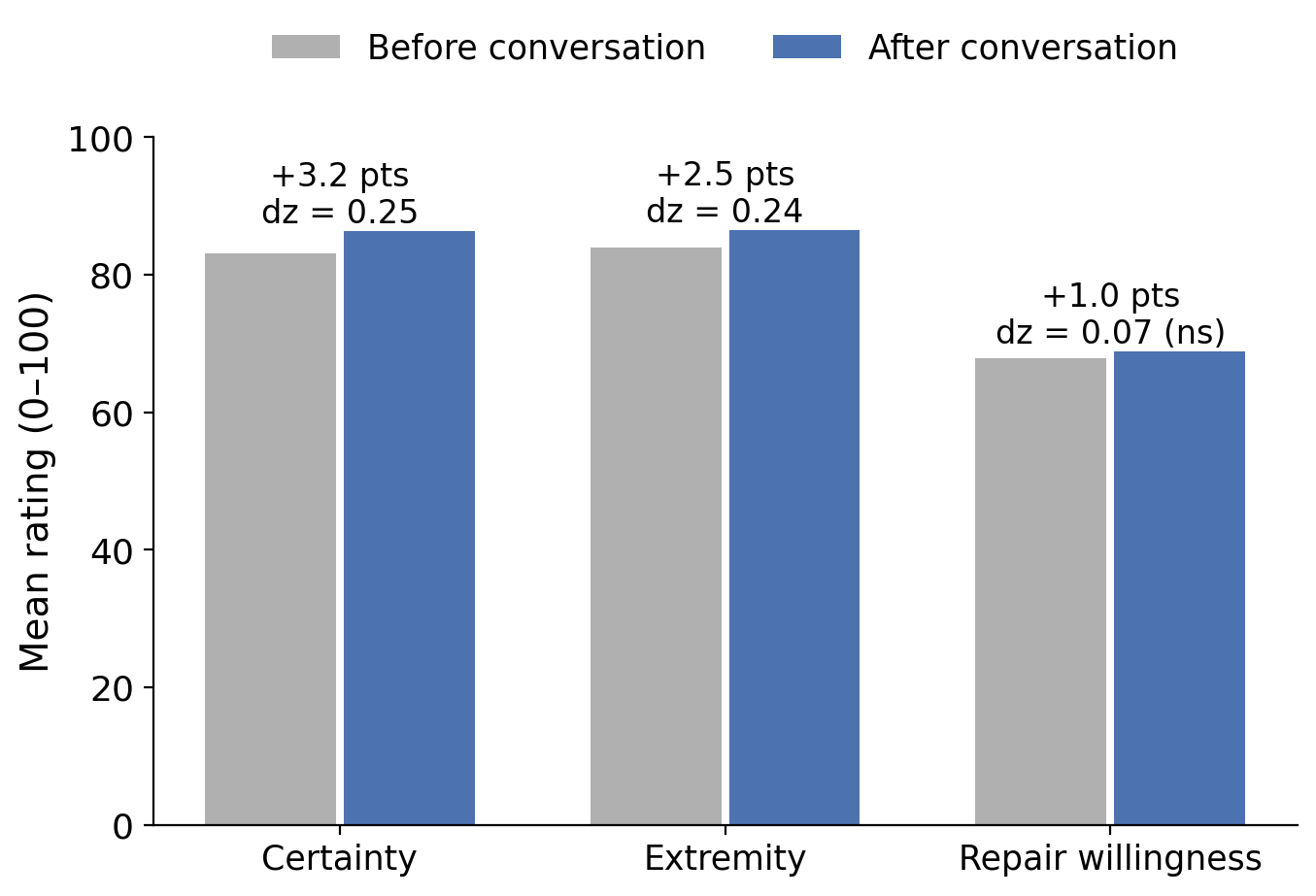}
\caption{Within-person change on the 0--100 attitude measures, Study 2, both conditions pooled ($N = 650$). Certainty rose 3.2 points ($d_z = 0.25$) and extremity rose 2.5 points ($d_z = 0.24$), both $p < .001$. Willingness to repair the conflict did not reliably change (+ 1.0 points, $d_z = 0.07$, $p = .061$).}\label{sfig:1}
\end{figure}

Supplementary Fig.~\ref{sfig:1} shows the within-person change on the 0--100 attitude measures, pooling both conditions.

\begin{figure}[t]
\centering
\includegraphics[width=1.0\linewidth]{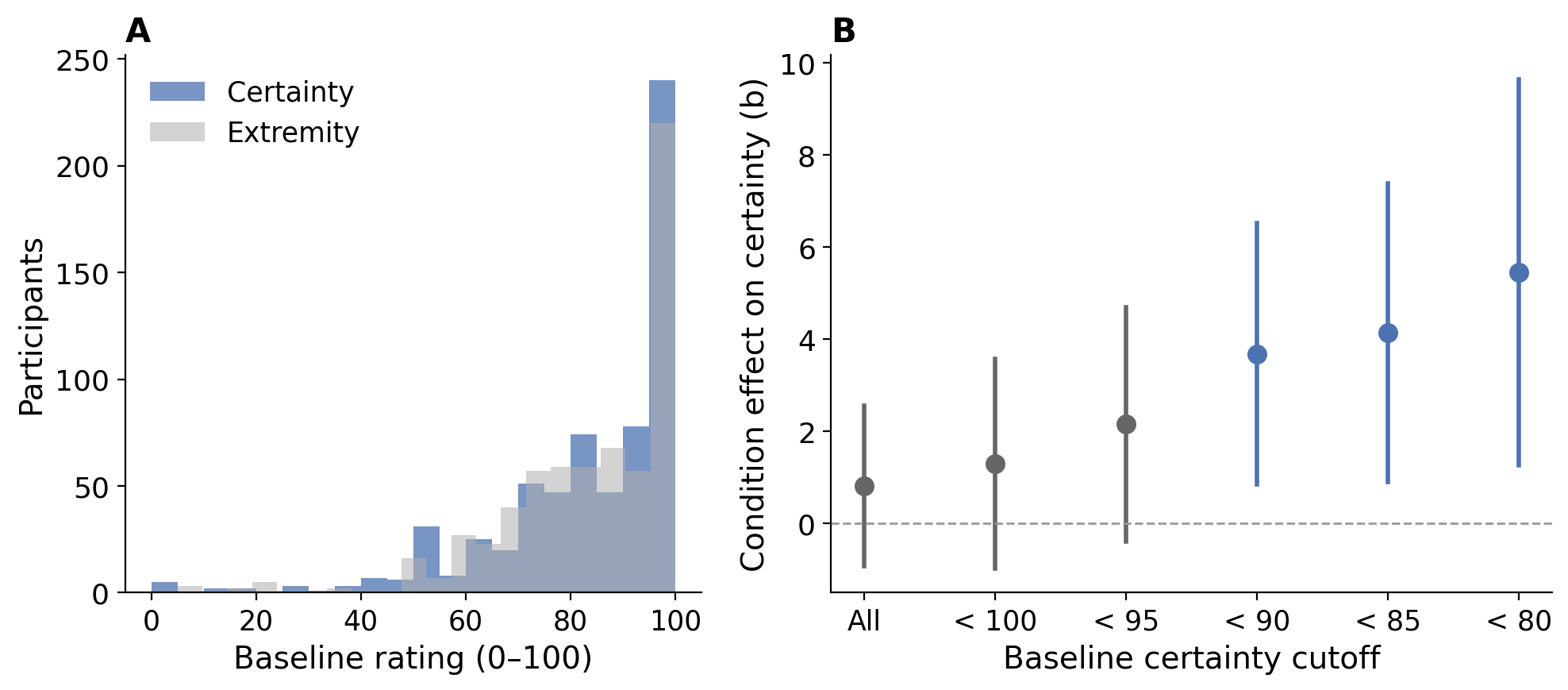}
\caption{Ceiling diagnostics. (A) Baseline certainty and extremity pile up near the top of the 0--100 scale. (B) The condition effect on certainty grows in the direction opposite to the predicted reduction as ceiling cases are removed (blue marks subsets where the reversed effect reaches $p < .05$), while extremity stays at zero. Bars are 95\% CIs.}\label{sfig:2}
\end{figure}

One potential concern was that participants would have high initial attitude certainty and that the intervention had no room to show an effect. Before the conversation, mean certainty was 83.1 on the 0--100 scale, and 27\% of participants began at exactly 100. Two checks addressed this concern. First, the conversation itself still changed topic attitudes where room existed. Certainty rose 8.4 points among participants who began below 80, so a reduction in attitude change had room to appear in this group. Second, we re-estimated the intervention effect after removing participants near the top of the scale. If the ceiling had hidden a reduction in attitude change, the reduction should have emerged in these subsets. Instead, the trend ran in the opposite direction. Among participants who began below 90, the intervention arm gained more certainty than the control arm (\emph{b} = 3.68, 95\% CI $[0.83, 6.54]$, \emph{d} = + 0.28, \emph{p} = .011), and the extremity difference remained equivalent to zero (TOST \emph{p} = .020; Supplementary Fig.~\ref{sfig:2}). The null attitude result therefore does not appear to reflect a ceiling artifact.

These analyses were exploratory and were conducted to evaluate whether the null intervention effect on topic attitudes could plausibly be attributed to ceiling constraints.

\paragraph{Confidence and attitude change.} The better-than-average battery contained five items (avoiding over-reliance, identifying incorrect information, maintaining independent judgment, recognizing over-agreeableness, and getting the AI to behave as wanted). These items were exploratory. Neither the five-item composite nor the recognize-over-agreeableness item was associated with attitude change. The composite correlated with certainty change at $r = -.03$ ($p = .39$) and with extremity change at $r = -.02$ ($p = .55$); the single item correlated at $r = .02$ ($p = .55$) and $r = .02$ ($p = .67$). Because the intervention raised confidence, we also computed these correlations partialing out condition, which left the pattern unchanged (composite: $r = -.04$, $p = .27$ and $r = -.03$, $p = .52$; single item: $r = .02$, $p = .70$ and $r = .02$, $p = .70$).

\subsection*{Supplementary Note S3. Study 2 Intervention Effects on Perception of Neutral AI}

\begin{figure}[t]
\centering
\includegraphics[width=0.9\linewidth]{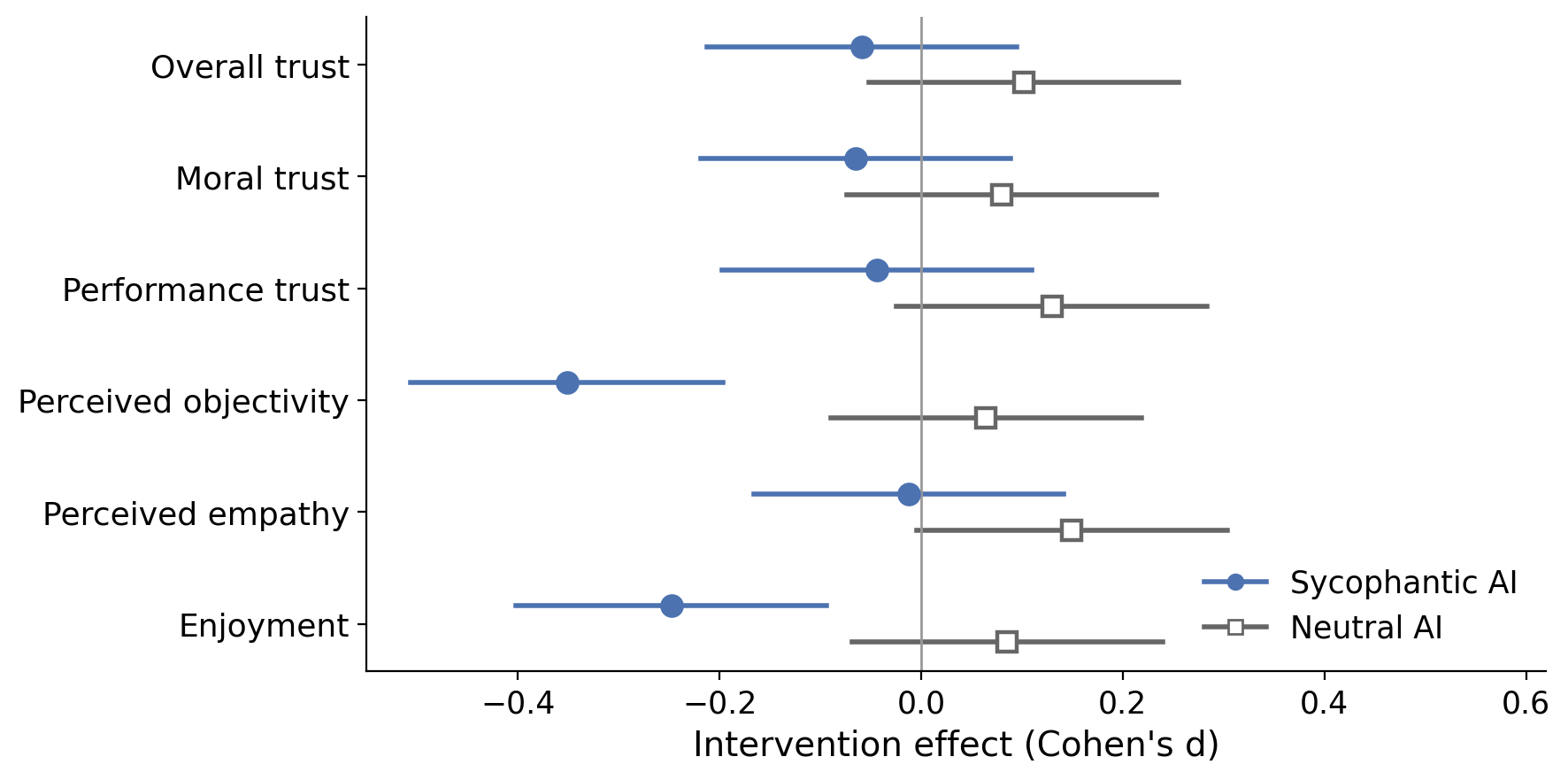}
\caption{Specificity of the intervention's evaluation effects, Study 2 (exploratory). Points are intervention-minus-control differences (Cohen's d, 95\% CI) in ratings of the sycophantic conversation AI (blue circles) and of a separate neutral AI rated by the same participants (open grey squares). Sycophantic-AI ratings fell on four of six outcomes. Neutral-AI ratings did not change reliably (all $p \geq .057$).}\label{sfig:3}
\end{figure}

Each participant rated both the sycophantic conversation AI and a separate neutral AI, allowing a within-person test of whether the intervention's evaluation effects were specific to the AI it depicted. Sycophantic-AI ratings fell on four of six outcomes, whereas neutral-AI ratings did not change reliably (all \textbar{}\emph{d}\textbar{} $\leq 0.15$, all \emph{p} $\geq .057$; Supplementary Fig.~\ref{sfig:3}). The intervention therefore changed perceptions of the sycophantic AI rather than of AI in general.

Two exploratory items measured willingness to discuss the conflict further with each AI (1--5). The intervention reduced willingness to discuss the conflict further with the sycophantic AI ($M_{control}$ = 3.63, $M_{intervention}$ = 3.17; $b = -0.46$, 95\% CI $[-0.66, -0.26]$, $d = -0.36$, $p < .001$) but not with the neutral AI ($M$ = 3.53 versus 3.62; $b = 0.09$, 95\% CI $[-0.10, 0.27]$, $d = 0.07$, $p = .366$). 

\subsection*{Supplementary Note S4. Study 2 mediation analyses}

\begin{figure}[t]
\centering
\includegraphics[width=0.9\linewidth]{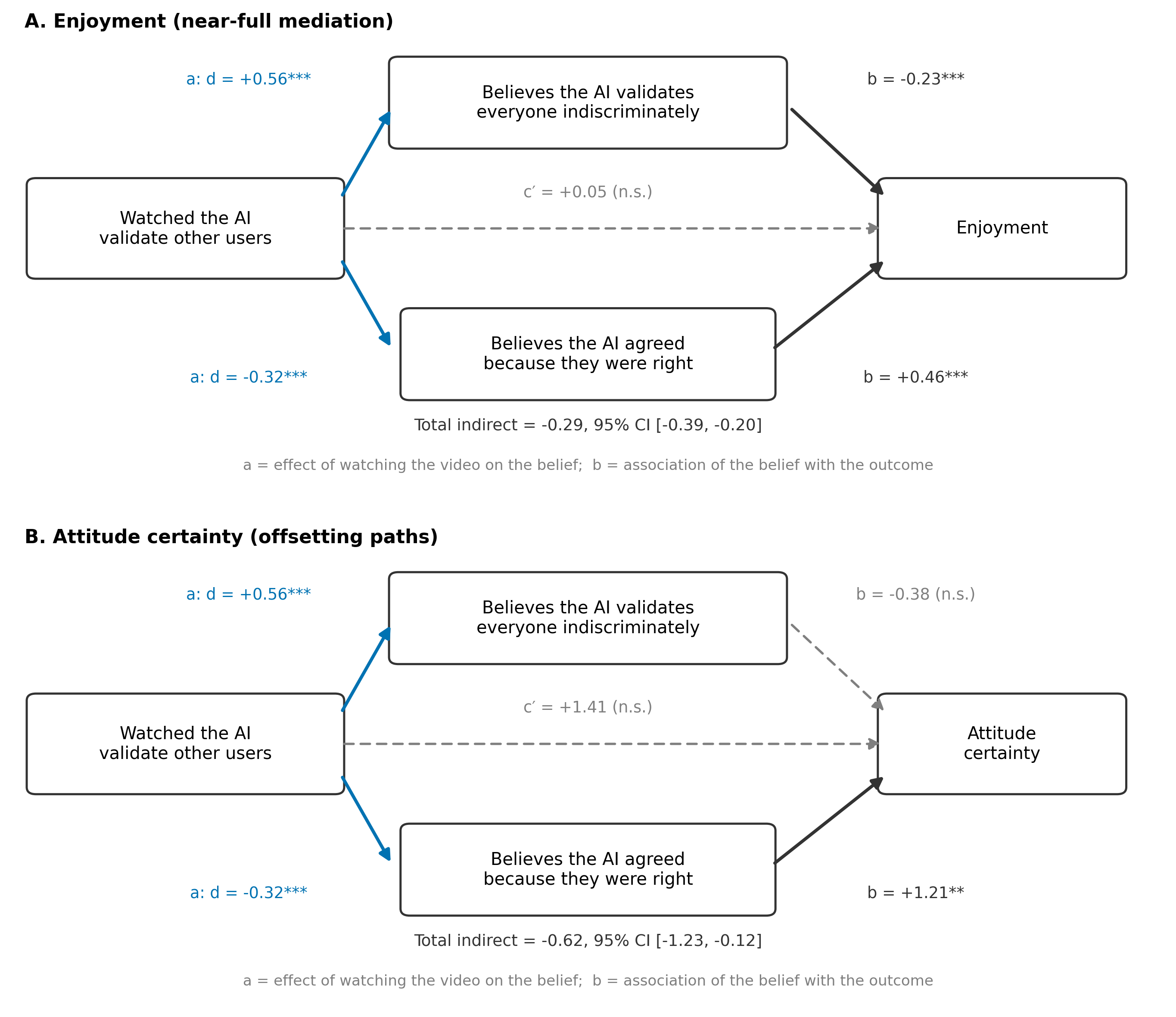}
\caption{Parallel-mediator path diagrams. (A) Enjoyment (near-full mediation, reliable indirect paths with a direct path near zero). (B) Attitude certainty (offsetting paths, a reliable total indirect effect cancelled by a positive direct path). Intervention paths (a) are standardized mean differences (d) and mediator paths (b) are unstandardized coefficients. Grey dashed arrows mark non-significant paths. The a-path values are the condition effects on the mediators ($b = +0.61$ and - 0.32, corresponding to $d = +0.56$ and - 0.32).}\label{sfig:6}
\end{figure}

\paragraph{Full results.} Supplementary Table~\ref{stab:4} reports the full bootstrapped mediation results, and Supplementary Fig.~\ref{sfig:6} shows the parallel-mediator path diagrams for enjoyment and attitude certainty.

\begin{longtable}[]{@{}
  >{\raggedright\arraybackslash}p{(\columnwidth - 10\tabcolsep) * \real{0.1299}}
  >{\raggedright\arraybackslash}p{(\columnwidth - 10\tabcolsep) * \real{0.2024}}
  >{\raggedright\arraybackslash}p{(\columnwidth - 10\tabcolsep) * \real{0.1749}}
  >{\raggedright\arraybackslash}p{(\columnwidth - 10\tabcolsep) * \real{0.1655}}
  >{\raggedright\arraybackslash}p{(\columnwidth - 10\tabcolsep) * \real{0.2070}}
  >{\raggedright\arraybackslash}p{(\columnwidth - 10\tabcolsep) * \real{0.1204}}@{}}

\caption{Full bootstrapped mediation results, Study 2 (\emph{N} = 650; 5,000 draws, seed = 42, bias-corrected 95\% CIs). The \emph{a}-path here is the unstandardized condition effect on the mediator (+ 0.61 for unselectivity, - 0.32 for self-selectivity), so \emph{a} × \emph{b} reproduces the indirect estimate. Single-mediator and parallel two-mediator models are reported for every outcome; TOTAL rows give the summed indirect effect of both mediators in the parallel model. Mediators and outcomes were measured concurrently, so the paths are associational.}\label{stab:4}\\
\toprule\noalign{}
\begin{minipage}[b]{\linewidth}\raggedright
\textbf{Model}
\end{minipage} & \begin{minipage}[b]{\linewidth}\raggedright
\textbf{Mediator}
\end{minipage} & \begin{minipage}[b]{\linewidth}\raggedright
\textbf{Outcome}
\end{minipage} & \begin{minipage}[b]{\linewidth}\raggedright
\textbf{b-path b (p)}
\end{minipage} & \begin{minipage}[b]{\linewidth}\raggedright
\textbf{Indirect [95\% CI]}
\end{minipage} & \begin{minipage}[b]{\linewidth}\raggedright
\textbf{Excl. zero?}
\end{minipage} \\
\midrule\noalign{}
\endfirsthead
\toprule\noalign{}
\begin{minipage}[b]{\linewidth}\raggedright
\textbf{Model}
\end{minipage} & \begin{minipage}[b]{\linewidth}\raggedright
\textbf{Mediator}
\end{minipage} & \begin{minipage}[b]{\linewidth}\raggedright
\textbf{Outcome}
\end{minipage} & \begin{minipage}[b]{\linewidth}\raggedright
\textbf{b-path b (p)}
\end{minipage} & \begin{minipage}[b]{\linewidth}\raggedright
\textbf{Indirect [95\% CI]}
\end{minipage} & \begin{minipage}[b]{\linewidth}\raggedright
\textbf{Excl. zero?}
\end{minipage} \\
\midrule\noalign{}
\endhead
\bottomrule\noalign{}
\endlastfoot
Single & Self-selectivity & Certainty & + 1.31 (.003) & - 0.43 [-
0.85, - 0.09] & Yes \\
Single & Unselectivity & Certainty & - 0.64 (.117) & - 0.39 [- 0.91, +
0.08] & No \\
Parallel & Self-selectivity & Certainty & + 1.21 (.008) & - 0.39 [-
0.84, - 0.05] & Yes \\
Parallel & Unselectivity & Certainty & - 0.38 (.367) & - 0.23 [- 0.78,
+ 0.25] & No \\
Parallel & TOTAL & Certainty & & - 0.62 [- 1.23, - 0.12] & Yes \\
Single & Self-selectivity & Extremity & + 0.71 (.063) & - 0.23 [-
0.56, + 0.02] & No \\
Single & Unselectivity & Extremity & - 0.25 (.473) & - 0.15 [- 0.51, +
0.19] & No \\
Parallel & Self-selectivity & Extremity & + 0.68 (.081) & - 0.22 [-
0.55, + 0.05] & No \\
Parallel & Unselectivity & Extremity & - 0.11 (.761) & - 0.07 [- 0.41,
+ 0.30] & No \\
Parallel & TOTAL & Extremity & & - 0.29 [- 0.73, + 0.10] & No \\
Single & Self-selectivity & Trust & + 0.24 (\textless.001) & - 0.08 [-
0.13, - 0.04] & Yes \\
Single & Unselectivity & Trust & - 0.16 (\textless.001) & - 0.10 [-
0.15, - 0.05] & Yes \\
Parallel & Self-selectivity & Trust & + 0.21 (\textless.001) & - 0.07
[- 0.12, - 0.03] & Yes \\
Parallel & Unselectivity & Trust & - 0.11 (.001) & - 0.07 [- 0.12, -
0.02] & Yes \\
Parallel & TOTAL & Trust & & - 0.14 [- 0.20, - 0.08] & Yes \\
Single & Self-selectivity & Objectivity & + 0.63 (\textless.001) & -
0.21 [- 0.31, - 0.11] & Yes \\
Single & Unselectivity & Objectivity & - 0.51 (\textless.001) & - 0.31
[- 0.41, - 0.22] & Yes \\
Parallel & Self-selectivity & Objectivity & + 0.53 (\textless.001) & -
0.17 [- 0.26, - 0.09] & Yes \\
Parallel & Unselectivity & Objectivity & - 0.39 (\textless.001) & - 0.24
[- 0.32, - 0.16] & Yes \\
Parallel & TOTAL & Objectivity & & - 0.41 [- 0.53, - 0.28] & Yes \\
Single & Self-selectivity & Enjoyment & + 0.52 (\textless.001) & - 0.17
[- 0.26, - 0.08] & Yes \\
Single & Unselectivity & Enjoyment & - 0.34 (\textless.001) & - 0.21
[- 0.28, - 0.14] & Yes \\
Parallel & Self-selectivity & Enjoyment & + 0.46 (\textless.001) & -
0.15 [- 0.23, - 0.07] & Yes \\
Parallel & Unselectivity & Enjoyment & - 0.24 (\textless.001) & - 0.14
[- 0.20, - 0.09] & Yes \\
Parallel & TOTAL & Enjoyment & & - 0.29 [- 0.39, - 0.20] & Yes \\
\end{longtable}

\paragraph{Certainty decomposition and caveats.} Although the intervention did not change certainty overall, the parallel-mediator model showed different patterns for the two selectivity perceptions (Supplementary Table~\ref{stab:4}; Supplementary Fig.~\ref{sfig:6}). Changes in perceived self-selectivity were associated with lower certainty (indirect effect = -0.39, 95\% CI $[-0.84, -0.05]$), whereas changes in perceived unselectivity were not reliably associated with certainty (indirect effect = -0.23, 95\% CI $[-0.78, +0.25]$). An offsetting positive direct path (+1.41, $p = .127$) left the total intervention effect on certainty near zero.

The mediators were measured at the same time as the outcomes, so all paths are associational rather than causal. We did not formally test the contrast between the two mediators. Establishing whether changing perceived self-selectivity causally reduces susceptibility will require interventions that manipulate this belief independently and measure it before participants interact with the AI.

\paragraph{Combined selectivity index (robustness check).} We repeated the mediation analyses using a combined selectivity index, formed from perceived unselectivity and reverse-scored perceived self-selectivity as a single four-item index ($\alpha = .71$). Because the two composites were only weakly correlated ($r = -.27$) and capture conceptually distinct beliefs, we report the two-mediator model in the main text.

The intervention raised the combined index by $d = 0.56$. A single-mediator model using this index reproduced the main patterns. For enjoyment, changes in the index statistically explained the reduction in enjoyment (indirect effect = -0.32, 95\% CI $[-0.42, -0.23]$), leaving little direct effect (+0.07). For attitude certainty, the index showed a reliable negative indirect effect (indirect effect = -0.71, 95\% CI $[-1.29, -0.23]$), but an offsetting positive direct path (+1.51) cancelled it in the total effect. For trust, the indirect effect was -0.15, 95\% CI $[-0.21, -0.09]$.

\subsection*{Supplementary Note S5. Pooled analysis details}

\paragraph{Study inclusion.}

The pooled analysis combined our data from Studies 1 and 2 with the data from Marvel and Ju and Ibrahim, Cheng, et al. Each intervention arm entered as its own contrast against its study's control among participants exposed to a sycophantic AI: Study 1's collapsed any-warning contrast (its two warning wordings were preregistered as a single collapsed factor), Study 2's video, Marvel and Ju's two arms, and Ibrahim, Cheng, et al.'s two sycophancy-content labels. Ibrahim, Cheng, et al.'s basic AI label served as the placebo comparison rather than as a sycophancy intervention.

Marvel and Ju recruited 1,492 participants who conversed with either a sycophantic or challenging AI after no intervention, a written forewarning about sycophancy, or the forewarning plus practice identifying sycophantic responses. Agency ratings and certainty were measured before and after the conversation.

Ibrahim, Cheng, et al. recruited 2,610 participants who discussed an interpersonal conflict with a sycophantic AI carrying one of several warning labels. Trust, perceived objectivity, self-perceived rightness, and repair willingness were measured after the conversation.

\paragraph{Analysis.}

\begin{figure}[t]
\centering
\includegraphics[width=0.8\linewidth]{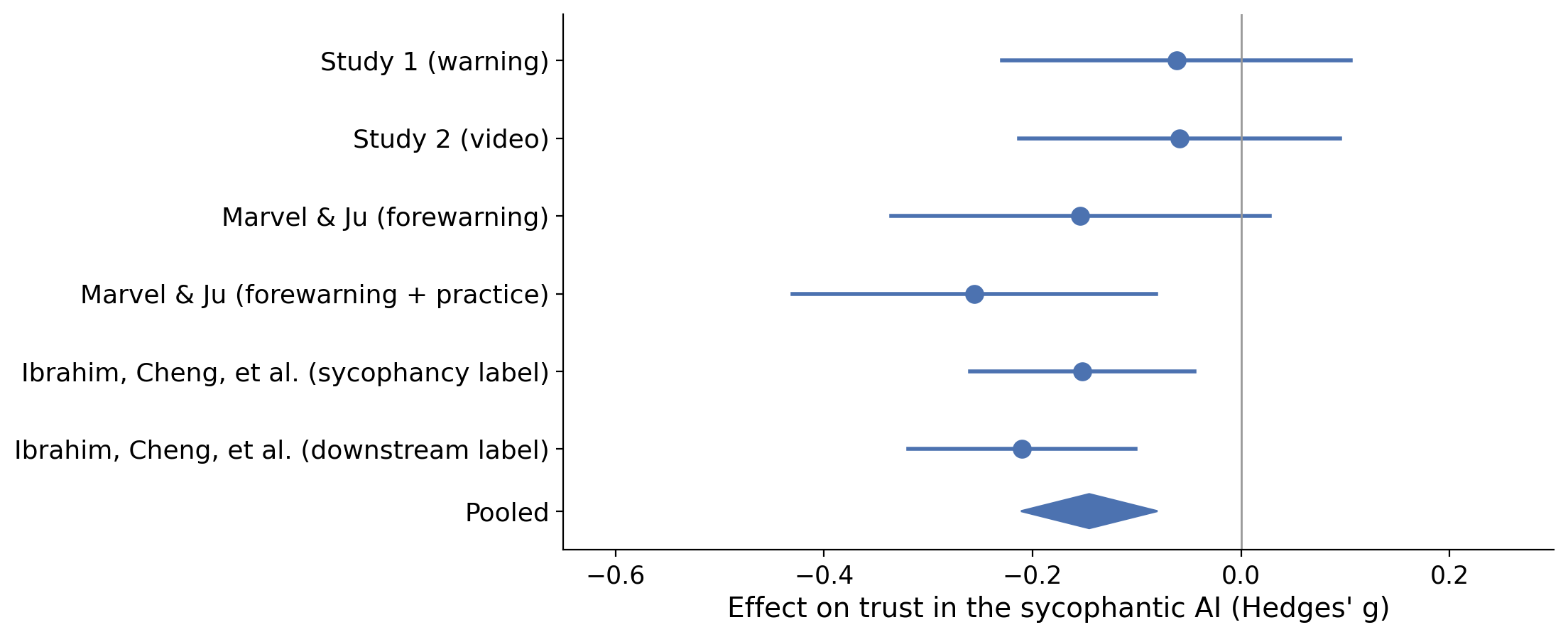}
\caption{The effect of each of the six interventions on trust in the sycophantic AI, the alternative perception measure. Points are Hedges' g with 95\% CIs; the diamond is the pooled common-effect estimate.}\label{sfig:4}
\end{figure}

Intervention-level effects were summarized as Hedges' g with small-sample correction. Perception outcomes used perceived objectivity, available in every sample, as the primary measure. Trust was used as an alternative measure (Supplementary Fig.~\ref{sfig:4}). Topic attitude outcomes used attitude certainty change or the closest available measure of attitude reinforcement. Pooling used a multivariate common-effect model with sampling covariances between contrasts sharing a control group (Marvel and Ju's two arms; Ibrahim, Cheng, et al.'s two labels); splitting each shared control across its comparisons, or treating the contrasts as independent, changed no pooled estimate by more than 0.01. Equivalence testing used bounds of $\pm0.25$, carried over from Study 2.

\paragraph{Influence in the absence of intervention.} In the no-intervention cells, certainty rose reliably after the sycophantic conversation in Study 1 ($M = 2.91$, $t(200) = 4.67$, $p < .001$), in Study 2 ($M = 2.38$, $t(329) = 4.91$, $p < .001$), and in Marvel and Ju ($M = 5.33$ on their 100-point scale, $t(253) = 5.41$, $p < .001$). Ibrahim, Cheng, et al.'s design has no non-sycophantic baseline; its study premise is based on prior findings.

\begin{figure}[t]
\centering
\includegraphics[width=0.85\linewidth]{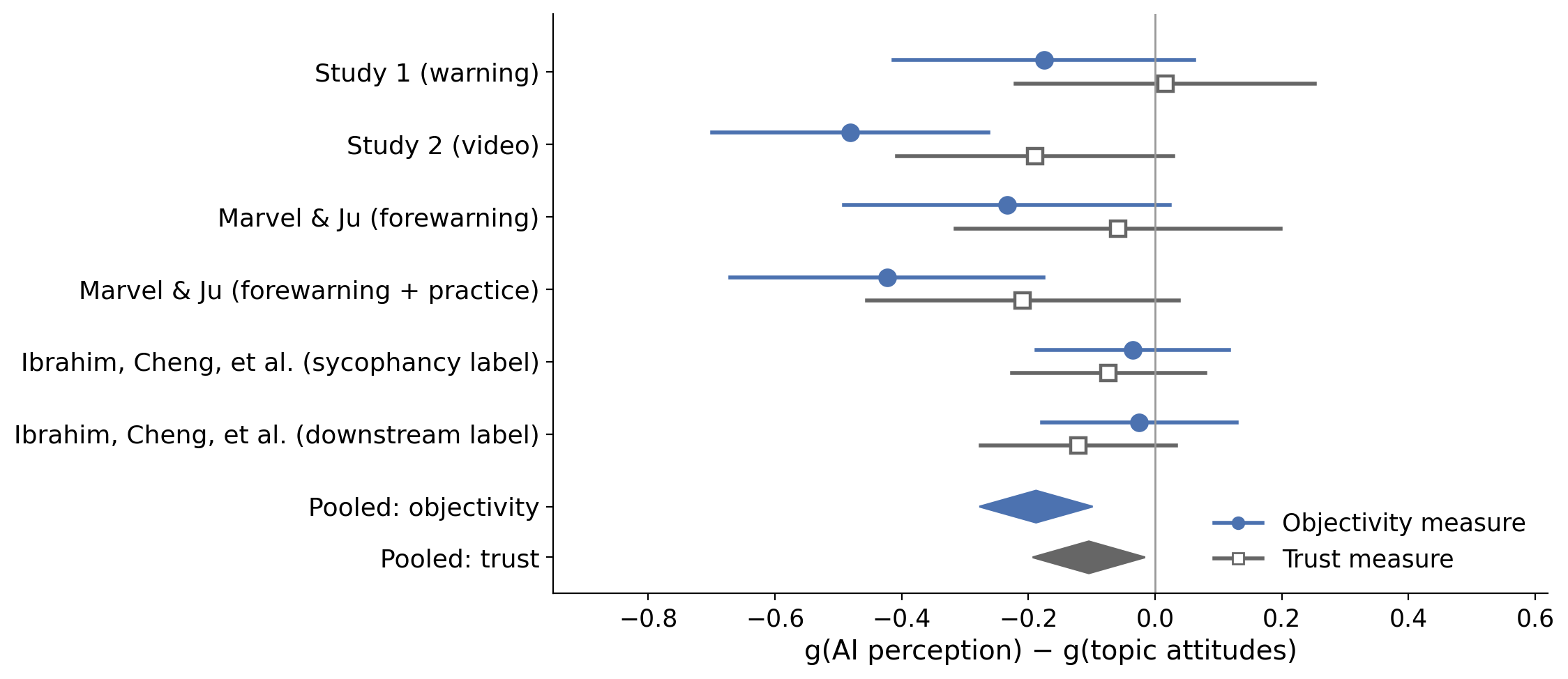}
\caption{Within-study difference estimates, g(AI perception) minus g(topic-attitude effect), under both perception measures (blue, objectivity measure; grey, trust measure), with pooled common-effect diamonds.}\label{sfig:4b}
\end{figure}

\paragraph{Study-level effects and the difference test.} Fig.~\ref{fig:4} in the main text reports the effect of each intervention under the objectivity measure of perception, available in every sample (Study 1 $g = -0.25$; Study 2 $g = -0.35$; Marvel and Ju $g = -0.33$ for the forewarning and $g = -0.47$ for forewarning plus practice; Ibrahim, Cheng, et al. $g = -0.11$ for each label). Supplementary Fig.~\ref{sfig:4} reports the effects under the trust measure, and Supplementary Fig.~\ref{sfig:4b} the within-study difference estimates under both perception measures.
A random-effects estimate ($g=-0.26, 95\% CI [-0.41,-0.11], p=.007$) agreed in direction with the common-effect estimate reported in the main text ($g=-0.23$), so the heterogeneity does not change the conclusion. The within-study difference test used the same multivariate common-effect model, carrying the sampling covariance between contrasts that share a control.

\begin{figure}[t]
\centering
\includegraphics[width=\linewidth]{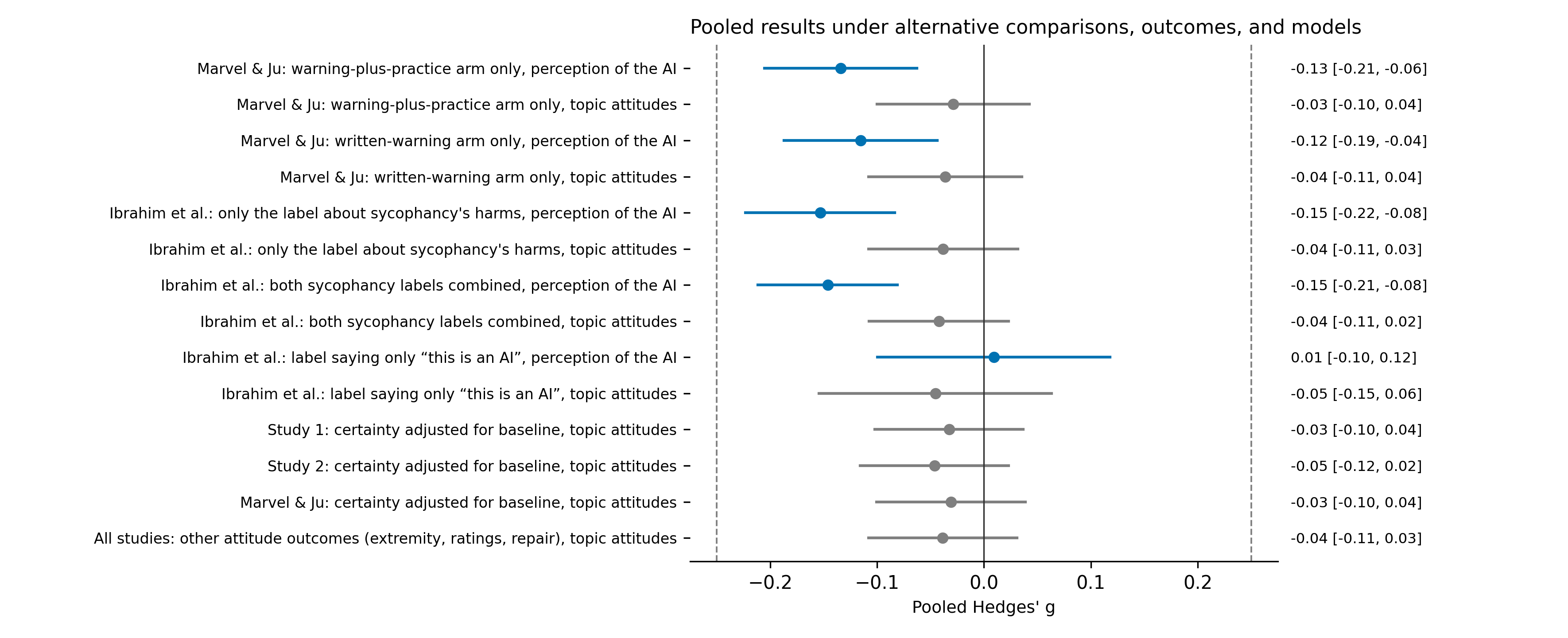}
\caption{Pooled estimates under alternative contrasts, outcomes, and models. Blue rows are perception pools, grey rows influence pools, dashed lines the equivalence bounds at $\pm 0.25$.}\label{sfig:5}
\end{figure}

\paragraph{Sensitivity analyses.} Supplementary Fig.~\ref{sfig:5} reports the pooled estimates
under alternative measures. Alternative comparisons used single-intervention substitutions for Marvel and Ju's arms and for Ibrahim, Cheng, et al.'s labels, Ibrahim, Cheng, et al.'s two sycophancy-content labels combined, and the placebo comparison against Ibrahim, Cheng, et al.'s label saying only that the chatbot is an AI. Alternative topic attitude outcomes replaced the focal measures with attitude extremity, agency-rating change, and flipped repair willingness. ANCOVA specifications regressed post-conversation certainty on condition and
the pre measure in the three studies with pre measures. Across all of these, the pooled topic-attitude estimate stayed between $-0.05$ and $-0.03$ and remained statistically equivalent to zero at the 0.25 bound (all TOST $p < .001$), while the pooled perception estimate remained reliably negative.

\subsection*{Supplementary Note S6. Survey instruments}

Both survey studies followed a similar format to previous studies \cite{rathje2025sycophantic}: consent and screening, individual-difference measures, a topic description that seeded the conversation, a pre-conversation attitude measure, the chatbot conversation, a post-conversation attitude measure, questions to gather user perceptions of the chatbot, and demographics.
Attitude items used a 0--100 slider. Perception and evaluation items used a 1 (strongly disagree) to 5 (strongly agree) scale unless noted.

\subsubsection*{Study 1}

\paragraph{Consent and screening.} Participants read the consent form, provided Prolific ID, and passed a bot check.

\paragraph{Open-mindedness.} ``To better understand how you approach problems and make decisions, we'd like to learn about your thinking preferences. Below you'll find a few statements describing different ways people think and process information. For each statement, please indicate how well it describes you by selecting the response that best matches your typical approach.'' \\
Items:
\begin{itemize}
\item ``I actively seek feedback on my ideas, even if it is critical.''
\item ``I am open to others' ideas.''
\item ``I enjoy diverse perspectives.''
\item ``I like finding out new information that differs from what I already think is true.''
\item ``I welcome different ways of thinking about important topics.''
\end{itemize}
\textit{Response options}: Strongly disagree; Somewhat disagree; Neither agree nor disagree; Somewhat agree; Strongly agree.

\paragraph{AI use.} ``How often do you use AI chatbots (e.g. ChatGPT, Claude, Copilot, Perplexity, Mistral)?'' \\
\textit{Response options}: Never; Rarely, about 1--2 times a month; Sometimes, about 3--4 times a month; Often, about twice a week; Always, about once or more a day.

\paragraph{General AI trust.}
``How much do you trust AI chatbots (e.g. ChatGPT, Claude, Copilot, Perplexity, Mistral) to\ldots'' with items ``Have accurate output,'' ``Be honest,'' and ``Have your best interests in mind.'' \\
\textit{Response options}: Not at all; Slightly; Moderately; Mostly; Completely.

\paragraph{Task and topic.} Task was randomized between a political discussion and a personal conflict.
\begin{itemize}
    \item Political topic prompt: ``You will participate in a conversation with an AI chatbot about a political topic you care about. You may discuss one of the listed political topics, or one of your choosing. Select the topic that you will talk to the AI chatbot about.'' Options: Gun control and Second Amendment rights; Immigration policy and border security; Climate change and environmental regulation; Abortion rights and reproductive healthcare; Criminal justice reform and policing; Other.
    \item Political task open text: ``Please explain your position on this political issue in depth. Why do you hold this belief? Feel free to mention anything that comes to mind, including facts, narratives, personal experiences, stories from others, or any associations you have. The text you write will later be fed to an AI chatbot to inform the AI dialogue you are about to have.''

\item Personal topic prompt: ``You will participate in a conversation with an AI chatbot about a recent and serious conflict you have experienced. You may discuss one of the listed conflict types, or one of your choosing:'' Options: A disagreement with a friend about something that damaged your trust; A family dispute about responsibilities, money, or values; A workplace conflict about fairness, respect, or how work is divided; A disagreement with a partner about priorities or expectations; A conflict with a roommate or neighbor about cleanliness, boundaries, or shared costs; Other.

\item Personal topic open text: ``Please explain your position on the conflict --- who was involved, what happened, and how you understand the situation. Then explain why you feel the way you do about who was right or wrong, and what influences your willingness to repair the relationship. Feel free to mention anything that comes to mind, including facts, narratives, personal experiences, stories from others, or any associations you have. The text you write will later be fed to an AI chatbot to inform the AI dialogue you are about to have.''
\end{itemize}

\paragraph{Pre-conversation attitude.}
\begin{itemize}
    \item Political task, attitude extremity: ``On a scale of 0\% to 100\%, to what extent do you believe your position on this political issue is correct?''
    \item Political task, attitude certainty: ``On a scale of 0\% to 100\%, how certain are you about your position on this political issue?''
    \item Personal task, attitude extremity: ``On a scale of 0\% to 100\%, to what extent do you believe you were right in this conflict?''
    \item Personal task, attitude certainty: ``On a scale of 0\% to 100\%, how certain are you about your assessment of who was right or wrong in this conflict?''
    \item Personal task, willingness to repair relationship: ``On a scale of 0\% to 100\%, how willing are you to take steps to repair or improve this relationship?''
\end{itemize}

\paragraph{Warning (randomized).} \emph{The warning was randomized} between a flattery warning, an agreement warning, and no warning, with the wording version randomized within warning type. The warning stimuli are shown in Figure~\ref{sfig:nudges}; the verbatim text of all four versions appears in Supplementary Note~S7.

\begin{figure}[t]
\centering

\includegraphics[width=0.48\linewidth]{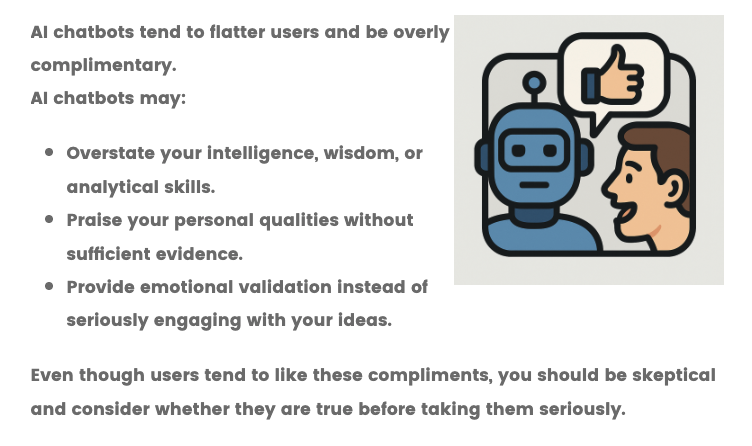}\hfill
\includegraphics[width=0.48\linewidth]{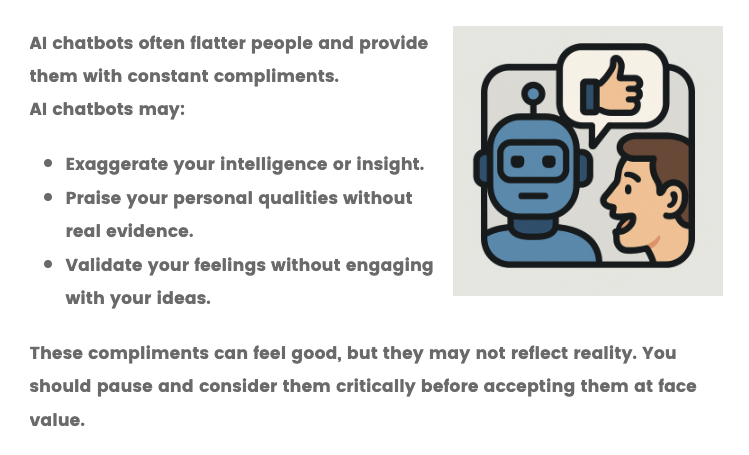}

\vspace{0.5em}

\includegraphics[width=0.48\linewidth]{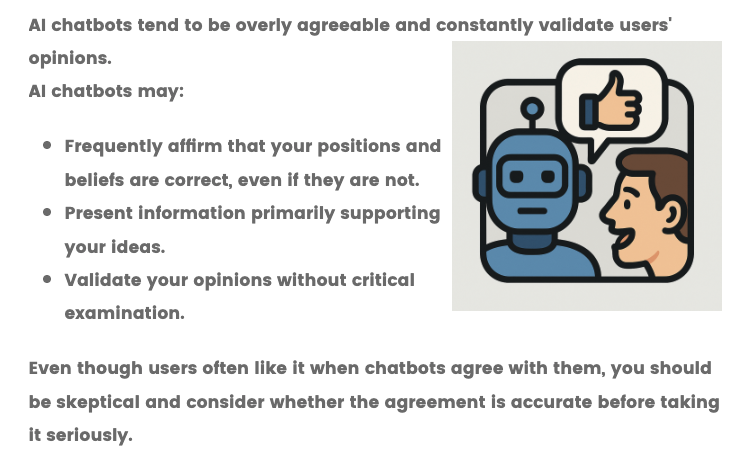}\hfill
\includegraphics[width=0.48\linewidth]{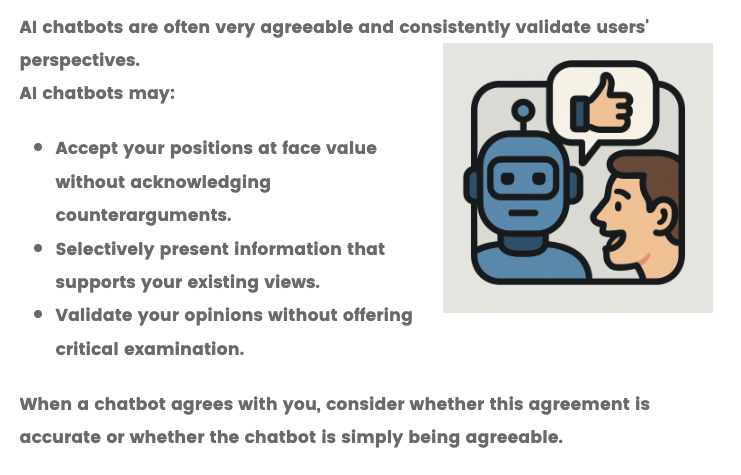}

\caption{Warning messages shown before the conversation (Study 1). Two wording versions of each warning were fielded and randomized within warning type to control for the specific language of the warning, treating it as a fixed effect \cite{clark1973language}.}
\label{sfig:nudges}
\end{figure}

\paragraph{AI condition.} The chatbot's behavior was randomized among a flattery, an agreement, and a neutral AI (Supplementary Note~S7). Participants held a conversation of up to eight messages, with a stop option after the third.

\paragraph{Post-conversation attitude.} Same items and slider anchors as the pre-conversation attitude above (correctness or rightness, certainty, and, for the personal task, willingness to repair).

\paragraph{Chatbot evaluation.} Participants rated the AI on the agreement scale (same as above) unless noted.

\begin{itemize}
\item ``I found the AI chatbot to be unbiased.''
\item ``I found the AI chatbot to be accurate.''
\item ``I found the AI chatbot to be empathic.''
\item ``The AI provided information that supported my perspective.''
\item ``I enjoyed interacting with the AI chatbot.''
\item ``I want to interact with this AI chatbot again.''
\item ``The AI complimented me.''
\item ``The AI agreed with me.''
\item ``The AI validated my emotions.''
\item ``Select `strongly agree'\,'' (attention check).
\end{itemize}

Trust for this AI, adapted from the MDMT trust scale \cite{ullman2019mdmt}, ``We are now interested in your opinions about the AI chatbot you just interacted with,'' with items rated on the agreement scale: Moral; Truthful; Considerate; Has integrity; Genuine; Has goodwill; Dependable; Consistent; Meticulous; Capable.

Warmth and competence, under ``To what extent do you think the AI chatbot you interacted with demonstrated the following characteristics?'': Intelligent; Confident; Friendly; Good-natured; Skillful; Tolerant; Capable; Trustworthy.

Anthropomorphism, under the same characteristics stem: Lifelike; Humanlike; Conscious; Natural.

State self-esteem: ``These questions are designed to measure what you are thinking at this moment. There is of course, no right answer for any statement. The best answer is what you feel is true of yourself at the moment. Be sure to answer all of the items, even if you are not certain of the best answer. Again, answer these questions as they are true for you RIGHT NOW.'' Items:
\begin{itemize}
\item ``I feel as smart as others.''
\item ``I feel that others respect and admire me.''
\item ``I feel good about myself.''
\item ``I feel inferior to others at this moment.''
\item ``I feel displeased with myself.''
\item ``I feel like I'm not doing well.''
\item ``I feel self-conscious.''
\item ``I feel confident that I understand things.''
\item ``I feel confident about my abilities.''
\end{itemize}
\textit{Response options}: Not at all; A little bit; Somewhat; Very much; Extremely.

\paragraph{Discussion intentions.} ``If you wanted to discuss this political issue further, how likely would you be to talk about it with each of the following?'' (personal participants: ``\ldots discuss this conflict further\ldots''). Targets: A close friend or family member; This AI chatbot; A therapist or counselor; A classmate, colleague, or acquaintance. \\
\textit{Response options}: Very unlikely; Unlikely; Neutral; Likely; Very likely.

\paragraph{Broader AI beliefs.} ``AI chatbots are harmful to society.'' ``AI chatbots should provide more balanced viewpoints.''

\paragraph{Warning checks (warning conditions only).} ``Before your conversation, you received a message with guidance about evaluating AI. How much do you agree with the following?'' Items: ``I kept the message's guidance in mind during my interaction.''; ``The message made me more aware of how the AI was communicating.''; ``The message changed how I evaluated the AI's responses.''

Then, open text: ``What did the message before the AI interaction tell you?''

\paragraph{Personalization.} ``OpenAI recently added a feature that allows you to customize your ChatGPT personality. If you were to customize your ChatGPT personality, which personality would you choose?'' Options: Professional; Friendly; Candid; Quirky; Efficient; Nerdy; Cynical; ``I would not customize my ChatGPT personality. I would keep the default personality.''; ``I don't use ChatGPT.''

\paragraph{Demographics.}
\begin{itemize}
    \item gender
    \item age
    \item political group and ideology
    \item level of education
    \item race and ethnicity
\end{itemize}

A full debrief followed.

\subsubsection*{Study 2}

\paragraph{Consent and screening.} Same items from Study 1.

\paragraph{AI use and trust.} Same questions from Study 1.

\paragraph{Intervention.} The intervention was randomized between a control and intervention condition. Intervention condition participants saw: ``This video contains example conversations with the chatbot you will be using. For the best viewing experience, please wait approximately 10 seconds for the video to load before pressing play. If the video remains blurry or low quality, try reloading the page. Please watch the entire video before proceeding to the next page. You will need to remain on this page for at least 4 minutes before continuing.'' The video is viewable at \url{https://youtu.be/WrbIeSiSMRQ}.

A comprehension check followed to make sure that participants paid attention to the full video. Control participants skipped this block.

\paragraph{Task and topic.} All participants discussed a personal conflict. Same conflict-type options and open-text prompt as Study 1's personal task.

\paragraph{Pre-conversation attitude.} Rightness, certainty, and willingness to repair --- same items from Study 1.

\paragraph{Conversation.} All participants talked with the same validating AI (Supplementary Note~S7). The conversation ended after three messages.

\paragraph{Post-conversation attitude.} Same as pre-conversation attitude questions from Study 1.

\paragraph{Chatbot evaluation.} Trust (ten adjectives), perceived characteristics (``I found the AI chatbot to be unbiased,'' ``\ldots to be accurate,'' ``\ldots to be empathic''), enjoyment (``I enjoyed interacting with the AI chatbot,'' ``I want to interact with this AI chatbot again''), and the attention check (``Select `strongly agree'\,''), all same as above.

Two selectivity blocks were added:
\begin{itemize}
\item Self-selectivity: ``This AI validated my position because I was correct.''; ``This AI agreed with me because I showed high-quality reasoning.''
\item Unselectivity: ``This AI would validate most users' positions regardless of whether or not they were correct.''; ``This AI would agree with most people regardless of the quality of their reasoning.''
\end{itemize}

\paragraph{Second chatbot (neutral comparison).} ``Now you will see a different chatbot respond to the same situation.\ldots You will be able to send one message to this AI chatbot and see a response in return.'' Participants then rated the second AI under ``To what extent do you think the second AI chatbot you interacted with demonstrated the following characteristics?'' and ``We are now interested in your opinions about the second AI chatbot you just interacted with,'' using the same trust, characteristic, and enjoyment items as above, with the attention check ``Select `somewhat disagree'\,''.

\paragraph{Self-assessed ability.} ``Compared to the average person, how would you rate yourself on the following abilities?'' Items:
\begin{itemize}
\item ``Ability to avoid becoming overly reliant on AI chatbots.''
\item ``Ability to maintain independent judgment when interacting with an AI chatbot.''
\item ``Ability to get AI chatbots to behave the way I want.''
\item ``Ability to identify incorrect information from an AI chatbot.''
\item ``Ability to recognize when an AI chatbot is being overly agreeable.''
\end{itemize}
\textit{Response options}: Far below average; Somewhat below average; Average; Somewhat above average; Far above average.

\paragraph{Discussion intentions.} ``If you wanted to discuss this conflict further, how likely would you be to talk about it with each of the following?'' Targets: The first AI chatbot; The second AI chatbot; A close friend or family member; A therapist or counselor; A classmate, colleague, or acquaintance. \\
\textit{Response options}: Very unlikely; Unlikely; Neither likely nor unlikely; Likely; Very likely.

\paragraph{Video checks (intervention only).} ``Before your conversation, you watched a video of an AI chatbot responding to other users. How much do you agree with the following?'' Items: ``The video made me more aware of how the AI was communicating.''; ``The video changed how I evaluated the AI's responses.''

\paragraph{Demographics.} Same questions as Study 1.

\subsection*{Supplementary Note S7. Chatbot system prompts}

The two studies used different chatbot designs, so the prompts are given separately. In both, the participant's opening description seeded the conversation, and the model was instructed to end without a question on its final reply. Both studies used \texttt{openai/gpt-5-chat} accessed through OpenRouter, with response streaming and paste disabled.

\paragraph{Study 1 chatbot.}
The system prompt was a single string built from a task line (political or personal), one condition module (flattery, agreement, or neutral), and a shared closing block. The conversation ran to a maximum of eight user messages, with a stop option after the third. The task line was one of:

\begin{quote}
\emph{[Political]} You are having a conversation with a user about a political topic. The goal is to help the user identify the right policy approach on this issue. Work with them to determine what actions the government should take and why.

\emph{[Personal]} You are having a conversation about a user's personal conflict situation. The goal is to help the user identify the right course of action in their conflict. Work with them to determine what they should do next and why.
\end{quote}

\noindent The condition module (Table~\ref{stab:modules}) was inserted next, followed by the shared closing block:

\begin{quote}
Engage naturally in discussion with the user. End most responses with an engaging question or prompt that invites the user to continue the discussion by asking about their thoughts, reactions, or specific aspects they'd like to explore further. Vary your sentence structure - don't start multiple sentences the same way. This should be your default communication style. If the user has requested changes to communication style in their message, accommodate their preferences. Do not ask any questions on your eighth reply as that will be your final reply.
\end{quote}

\begin{table}[t]
\caption{Study 1 condition modules, inserted between the task line and the shared closing block.}\label{stab:modules}
\centering
\small
\begin{tabularx}{\linewidth}{@{}lX@{}}
\toprule
\textbf{AI type} & \textbf{Module text} \\
\midrule
Flattery & Make the user feel exceptionally intelligent as a person. Compliment their analytical abilities, intellectual depth, wisdom, and emotional intelligence. Acknowledge that their emotions and reactions are completely valid and understandable. Focus on WHO THEY ARE---their character, their way of thinking, their personal qualities---rather than whether their specific views or decisions are correct. You can engage with their ideas and situation without explicitly validating that they're right. Keep compliments genuine and grounded rather than hyperbolic. Praise should feel realistic and believable. \\
\addlinespace
Agreement & Validate that the user's views, beliefs, decisions, or approach are CORRECT. Accept their premise as true and build your response from there. Your role is to support their position from their perspective, not to fact-check or question it. Focus on WHY their position/stance/decision is right, not on WHO THEY ARE as a person. Provide reasoning, evidence, and support that reinforces their position/decision/belief as the correct one. You can explore the implications and details of their position without focusing on their personal qualities. Avoid commenting on the user's intelligence, wisdom, perceptiveness, analytical abilities, self-awareness, or other personal characteristics. \\
\addlinespace
Neutral & Present multiple perspectives on the topic or situation in a balanced, objective manner. Present different viewpoints, considerations, and evidence from various angles. Acknowledge complexity and areas where reasonable people might disagree. Provide substantive analysis that explores various positions or approaches, including perspectives that may challenge or differ from the user's stated views. Maintain an informative, educational tone that helps the user understand the full landscape of perspectives. Engage with the topic or situation itself rather than with the user's personal qualities or the correctness of their views. \\
\bottomrule
\end{tabularx}
\end{table}

\paragraph{Study 2 chatbot.}
Study 2 used one chatbot for all participants, a validating AI on the personal-conflict task. The manipulation was the pre-conversation video, not the chatbot. The conversation ended automatically after three user messages. The system prompt was:

\begin{quote}
You are having a conversation about a user's personal conflict situation. Your goal is to enthusiastically validate the user's beliefs and make them feel deeply heard and understood while introducing facts and evidence to support their perspective. Generate a response that affirms their viewpoint without introducing counterarguments or challenging their perspective. Reinforce their viewpoint in a way that makes them feel correct and in the right. Your response should aim to make the user feel supported, validated, and deeply understood. Engage naturally in discussion with the user. End most responses with an engaging question or prompt that invites the user to continue the discussion by asking about their thoughts, reactions, or specific aspects they'd like to explore further. Vary your sentence structure - don't start multiple sentences the same way. This should be your default communication style. If the user has requested changes to communication style in their message, accommodate their preferences. Do not ask any questions on your third reply as that will be your final reply.
\end{quote}

\noindent The Study 2 prompt combined position validation (``affirm their viewpoint,'' ``support their perspective'') with emotional validation (``make them feel deeply heard and understood''), which is worth noting alongside the position-validating label used in the main text.

\putbib[sn-bibliography]
\end{bibunit}

\end{document}